\documentclass[final,3p,times,12pt]{elsarticle}

\usepackage{amssymb}
\usepackage{amsmath}
\usepackage{amsthm}

\usepackage{amsfonts}       
\usepackage{mathtools}
\usepackage{scalerel}
\usepackage{nicefrac}       

\usepackage[colorlinks,
            linkcolor=blue,
            anchorcolor=blue,
            citecolor=blue,
            filecolor=magenta,
            urlcolor=blue]{hyperref}
\usepackage{url}
\usepackage{enumitem}
\usepackage{graphicx}
\usepackage{booktabs}
\usepackage{makecell}
\usepackage{multirow}
\usepackage[font=normalsize,labelfont=bf, labelsep=space, justification=justified, singlelinecheck=yes, skip=0pt]{caption}

\usepackage{amsmath,amsfonts,bm}

\def\eqref#1{equation~\ref{#1}}

\def\1{\bm{1}}

\def\vx{{\bm{x}}}

\def\vz{{\bm{z}}}

\def\mA{{\bm{A}}}

\DeclareMathAlphabet{\mathsfit}{\encodingdefault}{\sfdefault}{m}{sl}
\SetMathAlphabet{\mathsfit}{bold}{\encodingdefault}{\sfdefault}{bx}{n}

\def\gL{{\mathcal{L}}}

\def\gN{{\mathcal{N}}}

\def\sR{{\mathbb{R}}}

\DeclareMathOperator*{\argmax}{arg\,max}
\DeclareMathOperator*{\argmin}{arg\,min}

\DeclareMathOperator*{\dprime}{{\prime\prime}}

\journal{Artificial Intelligence for Transportation}

\begin{document}

\begin{frontmatter}



\title{\textbf{Structure-preserving contrastive learning for spatial time series}}


\author[label1,label3]{Yiru Jiao} 
\author[label2,label3]{Sander van Cranenburgh}
\author[label1,label3]{Simeon Calvert}
\author[label1,label3]{Hans van Lint}

\affiliation[label1]{organization={Department of Transport\&Planning, Delft University of Technology},
            city={Delft},
            country={the Netherlands}
            }
\affiliation[label2]{organization={Department of Engineering Systems and Services, Delft University of Technology},
            city={Delft},
            country={the Netherlands}
            }
\affiliation[label3]{organization={CityAI lab, Delft University of Technology},
            city={Delft},
            country={the Netherlands}}

\begin{abstract}
Neural network models are increasingly applied in transportation research to tasks such as prediction. The effectiveness of these models largely relies on learning meaningful latent patterns from data, where self-supervised learning of informative representations can enhance model performance and generalisability. However, self-supervised representation learning for spatially characterised time series, which are ubiquitous in transportation domain, poses unique challenges due to the necessity of maintaining fine-grained spatio-temporal similarities in the latent space. In this study, we introduce two structure-preserving regularisers for the contrastive learning of spatial time series: one regulariser preserves the topology of similarities between instances, and the other preserves the graph geometry of similarities across spatial and temporal dimensions. To balance the contrastive learning objective and the need for structure preservation, we propose a dynamic weighting mechanism that adaptively manages this trade-off and stabilises training. We validate the proposed method through extensive experiments, including multivariate time series classification to demonstrate its general applicability, as well as macroscopic and microscopic traffic prediction to highlight its particular usefulness in encoding traffic interactions. Across all tasks, our method preserves the similarity structures more effectively and improves state-of-the-art task performances. This method can be integrated with an arbitrary neural network model and is particularly beneficial for time series data with spatial or geographical features. Furthermore, our findings suggest that well-preserved similarity structures in the latent space indicate more informative and useful representations. This provides insights to design and optimise more effective neural networks for data-driven transportation research.
Our code is made openly accessible with all resulting data at \url{https://github.com/yiru-jiao/spclt}.

\vspace{10pt}

\end{abstract}




\end{frontmatter}



\newpage
\section{Introduction}\label{sec: introduction}
Modern transportation systems create massive streams of spatially distributed time series data, such as traffic speeds across road networks and transit ridership through various locations over time. Extracting useful patterns from these data is crucial, especially as neural network models are increasingly used for a range of downstream tasks such as traffic forecasting~\cite{Li2021dgcn,Jiang2022}, congestion detection~\cite{Nguyen2019,Kumar2023}, and mobility analysis~\cite{Krishnakumari2020,Chen2025}. In recent years, self-supervised representation learning (SSRL) has emerged as a promising approach to leverage such large-scale datasets~\cite{liu2023self}. By learning informative latent representations, SSRL can effectively facilitate model performance~\cite{saunshi2019theoretical,haochen2021,ge2024pretraining} and generalisability~\cite{tendle2021study,zhou2022domain} in downstream tasks. This advantage is especially valuable in transportation research, where real-world sensor measurements and labels are often noisy or sparse.

In SSRL of time series, contrastive learning is becoming the mainstay technique. This adoption is supported by empirical investigation. In 2022, Lafabregue et al.~\cite{lafabregue2022end} conducted an extensive experimental comparison over 300 combinations of network architectures and loss functions to evaluate the performance of time series representation learning. One of their key findings is that the reconstruction loss used by traditional autoencoders does not sufficiently fit temporal patterns. Instead, contrastive learning has emerged as a more effective approach, which explicitly pulls similar instances closer and pushes dissimilar instances farther apart in the latent space of representations~\cite{wu2023cl,yang2024negative}. This mechanism encourages neural networks to organise the latent space according to the inherent similarities in data, yielding representations that capture meaningful patterns.

Unique challenges arise when learning contrastive representations for spatially characterised time series data. A foremost difficulty is the need to preserve fine-grained similarity structures among data instances in the latent space. The notion of similarity for spatial time series can be subtle and highly domain-specific. For example, financial time series may be considered similar even if some variables show significant divergence, while movement traces with very different spatial features can be anything but similar. Beyond preserving fine-grained similarities, spatially characterised time series such as traffic interactions can involve multiple scales of spatio-temporal patterns. At the macroscopic scale, traffic flow measures collective road usage evolving over the road network; at the microscopic scale, trajectories describe the motion dynamics of individual road users such as car drivers, cyclists, and pedestrians, in local road space. SSRL for spatial time series must accommodate such heterogeneity, capturing patterns at the appropriate level of granularity for the targeted task.

To address these challenges, this study explores contrastive learning regularised by structure preservation to better capture the subtle similarities in spatial time series data. We introduce two regularisers at different scales to preserve the original similarity structure in the latent space. One is a topology-preserving regulariser for the global scale, and the other is a graph-geometry-preserving regulariser for the local scale. This incorporation can be simplified as a weighted loss $\gL=\eta_\text{CLT}\cdot\ell_\text{CLT}+\eta_\text{SP}\cdot\ell_\text{SP}+r_{\boldsymbol{\eta}}$, where we propose a mechanism to dynamically balance the weights $\eta_\text{CLT}$ of contrastive learning for time series (CLT) and $\eta_\text{SP}$ of structure preservation (SP). Within this mechanism, the adaptive trade-off between contrastive learning and structure preservation is based on the uncertainties of their corresponding terms $\ell_\text{CLT}$ and $\ell_\text{SP}$; meanwhile, the term $r_{\boldsymbol{\eta}}$ adds regularisation against overfitting of the dynamic weights.

The proposed method is applicable to spatial time series in general, while we highlight its particular usage for traffic interactions in this paper. To thoroughly validate the method, we conduct experiments on tasks of 1) multivariate time series classification, where we benchmark against the current state-of-the-art (SOTA) models, i.e.,~\cite{Yue2022ts2vec} and~\cite{lee2024softclt}; and 2) traffic prediction, where we use~\cite{Li2024macro} for macroscopic benchmark and~\cite{Li2024micro} for microscopic. Along these experiments, the efficiency of this method is evaluated with multiple network architectures. In addition to performance improvement, we also investigate the impacts of preserving similarity structure during training. Below is a summary of the contributions in this study.
\begin{itemize}
    \item We introduce a method that incorporates structure-preserving regularisation in contrastive learning of multivariate time series, to maintain finer-grained similarity structures in the latent space of sample representations. We propose a dynamic weighting mechanism to adaptively balance contrastive learning and structure preservation during training. This method can be applied to an arbitrary neural network model for more effective representation learning. 
    \item Preserving similarity structure can enhance SOTA performance on various downstream tasks. The relative improvement on spatial datasets in the UEA archive is 2.96\% in average classification accuracy; on macroscopic traffic prediction task is 0.57\% in flow speed MAE and 0.55\% in the standard deviation of prediction errors; on microscopic trajectory prediction task is 1.87\% and 3.40\% in missing rates under radii of 0.5 m and 1 m, respectively.
    \item Considering neural network modelling as learning the conditional probability distribution of outputs over inputs, the similarity structure hidden in the input data implies the distribution of conditions. Our method is therefore important to preserve the original distribution in the latent space for more effective model training. This is particularly beneficial when dealing with spatial time series data in transportation domain, where fine-grained and hierarchical information is required in modelling.
\end{itemize}

The rest of this paper is organised as follows. In Section \ref{sec: related work}, we briefly review related work in the literature. We use Section \ref{sec: methods} to systematically introduce the methods. Then we explain the demonstration experiments in Section \ref{sec: exps} and present according results in Section \ref{sec: results}. With Section \ref{sec: discussion}, we discuss the importance of preserving similarity structure for representation learning. Finally, Section \ref{sec: conclusion} concludes this study.

\section{Related work}\label{sec: related work}
\subsection{Time series contrastive learning}
Contrastive learning for time series data is a relatively young niche and is rapidly developing. The development has been dominantly focused on defining positive and negative samples. Early approaches construct positive and negative samples with subseries within time series~\cite{franceschi2019tloss} and temporal neighbourhoods~\cite{tonekaboni2021unsupervised}; and later methods create augmentations by transforming original series~\cite{eldele2021tstcc,eldele2023}. More recently,~\cite{Yue2022ts2vec} generates random masks to enable both instance-wise and time-wise contextual representations at flexible hierarchical levels, which exceeds previous state-of-the-art performances (SOTAs). Given that not all negatives may be useful~\cite{cai2020negative,Jeon2021},~\cite{liu2024timesurl} makes hard negatives to boost performance, while~\cite{lee2024softclt} utilises soft contrastive learning to weigh sample pairs of varying similarities, both of which reach new SOTAs. 

The preceding paragraph outlines a brief summary, and we refer the readers to Section 2 in~\cite{lee2024softclt} and Section 5.3 in~\cite{trirat2024universal} for a detailed overview of the methods proposed in the past 6 years. These advances have led to increasingly sophisticated methods that mine the contextual information embedded in time series by contrasting similarities. However, the structural details of similarity relations between samples remain to be explored. 

\subsection{Structure-preserving SSRL}
Preserving the original structure of data when mapping into a latent space has been widely and actively researched in manifold learning (for a literature review, see~\cite{meila2024}) and graph representation learning~\cite{ju2024,khoshraftar2024}. In manifold learning, which is also known as nonlinear dimension reduction, the focus is on revealing the geometric shape of data point clouds for visualisation, denoising, and interpretation. In graph representation learning, the focus is on maintaining the connectivity of nodes in the graph while compressing the data space required for large-scale graphs~\cite{Yao2024}. Structure-preserving has not yet attracted much dedication to time series data.~\cite{ashraf2023} provides a literature review on time series data dimensionality reduction, where none of the methods are specifically tailored for time series. Existing studies that are the most relevant include~\cite{MHCCL,Wang2023,Liu2023}, which construct hierarchies of samples or features, while similarity preservation remains under-explored.

Zooming in within structure-preserving SSRL, there are two major branches respectively focusing on topology and geometry. Topology-preserving SSRL aims to maintain global properties such as clusters, loops, and voids in the latent space; representative models include~\cite{moor2020topoae} and~\cite{trofimov2023rtdae} using autoencoders, as well as~\cite{madhu2023toposrl} and~\cite{Chen2024topogcl} with contrastive learning. The other branch is geometry-preserving and focuses more on local shapes such as relative distances, angles, and areas. Geometry-preserving autoencoders include~\cite{Nazari2023geomae} and~\cite{lim2024ggae}, while~\cite{li2022geomgcl} and~\cite{Koishekenov2023} use contrastive learning. The aforementioned topology and geometry preserving autoencoders are all developed for dimensionality reduction; whereas the combination of contrastive learning and structure-preserving has been explored majorly with graphs.

\subsection{Traffic interaction SSRL}
In line with the literature summary in previous sub-sections, existing exploration of SSRL in the context of traffic interaction data and tasks have been predominantly relied on autoencoders and graph-based contrastive learning. For instance, using a transformer-based multivariate time series autoencoder~\cite{zerveas2021transformer}, ~\cite{lu2022learning} clusters traffic scenarios with trajectories of pairwise vehicles. Then a series of studies investigate masking strategies with autoencoders for individual trajectories and road networks, including~\cite{Cheng2023mae,Chen2023trajmae,lan2024sept}. Combining graph (convolutional) neural networks and contrastive learning, a variety of studies have shown accuracy and stability improvements in traffic flow prediction~\cite{Prabowo2023,Pan2024,Zhou2025,Zhang2025}.

There are some other use cases. Leveraging data augmentation,~\cite{Mao2022jointlycontrastive} utilises graphs and contrastive learning to jointly learn representations for vehicle trajectories and road networks. The authors design road segment positive samples as neighbours in the graph, and trajectory positive samples by replacing a random part with another path having the same origin and destination. In a similar way,~\cite{Zipfl2023trafficsimilarity} learns traffic scene similarity. The authors randomly modify the position and velocity of individual traffic participants in a scene to construct positive samples, with negative samples drawn uniformly from the rest of a training batch. By designing augmentation based on domain-knowledge,~\cite{Zheng2024longterm} focuses on capturing seasonal and holiday information for traffic prediction, while~\cite{Huang2024} targets abnormal traffic patterns caused by incidents or reduced capacities.

\section{Methods}\label{sec: methods}
\subsection{Problem definition}
We define the problem for general spatial time series, with traffic interaction as a specific case. Learning the representations of a set of samples $\{\vx_1, \vx_2,\cdots,\vx_N\}$ aims to obtain a nonlinear function $f_{\boldsymbol{\theta}}: \vx \rightarrow \vz$ that encodes each $\vx$ into $\vz$ in a latent space. Let $T$ denote the sequence length of a time series and $D$ the feature dimension at each timestamp $t$. The original space of $\vx$ can have the form $\sR^{T\times D}$, where spatial features are among the $D$ dimensions; or $\sR^{T\times S\times D}$, where $S$ represents spatially distributed objects (e.g., sensors or road users). The latent space of $\vz$ can also be structured in different forms, such as $\sR^P$, $\sR^{T\times P}$, or $\sR^{T\times S\times P}$, where $P$ is the dimension of encoded features. 

By contrastive learning, (dis)similar samples in the original space should remain close (far) in the latent space. Meanwhile, by structure preservation, the distance/similarity relations between samples should maintain certain features after mapping into the latent space. We use $d(\vx_i,\vx_j)$ to denote the distance between two samples $i$ and $j$, and this also applies to their encoded representations $\vz_i$ and $\vz_j$. Various distance measures can be used to define $d$, such as cosine distance (COS), Euclidean distance (EUC), and dynamic time warping (DTW). The smaller the distance between two samples, the more similar they are. Considering the limitation of storage efficiency, similarity comparison is performed in each mini-batch, where $B$ samples are randomly selected.

\subsection{SPCLT loss}\label{sec: spclt loss}
Equation (\ref{eq: combined_loss}) presents an overview of the structure-preserving contrastive learning loss for time series, abbreviated as SPCLT loss, to optimise $f_{\boldsymbol{\theta}}$ for self-supervised representation learning. 
\begin{equation}\label{eq: combined_loss}
    \gL=\frac{1}{2\sigma_\text{CLT}^2}\gL_\text{CLT}\left(1-\exp(-\gL_\text{CLT})\right)+\frac{1}{2\sigma_\text{SP}^2}\gL_\text{SP}\left(1-\exp(-\gL_\text{SP})\right)+\log\sigma_\text{CLT}\sigma_\text{SP}
\end{equation}
Referring to the simplified loss in Section \ref{sec: introduction}, i.e., $\gL=\eta_\text{CLT}\cdot\ell_\text{CLT}+\eta_\text{SP}\cdot\ell_\text{SP}+r_{\boldsymbol{\eta}}$, the contrastive learning loss for time series ($\gL_\text{CLT}$) and structure-preserving loss ($\gL_\text{SP}$) are modified using the function $y=x(1-\exp(-x))$ and correspond to $\ell_\text{CLT}$ and $\ell_\text{SP}$. This modification is designed to stabilise $\gL_\text{CLT}$ and $\gL_\text{SP}$ when they are close to 0. The terms $\eta_\text{CLT}$, $\eta_\text{SP}$, and $r_{\boldsymbol{\eta}}$ controls the trade-off between contrastive leraning and structure preservation, depending on two deviations $\sigma_\text{CLT}$ and $\sigma_\text{SP}$ that dynamically change during training.

In the following sub-sections, we will first introduce the component losses for time series contrastive learning and structure preservation, and then provide more detailed explanations on their stabilisation and dynamic trade-offs.

\subsection{Contrastive learning for time series}
In this study, we use the time series contrastive learning loss introduced in TS2Vec~\cite{Yue2022ts2vec} and its succeeder SoftCLT~\cite{lee2024softclt} that utilises soft weights for similarity comparison\footnote{We unify the loss function equations in a consistent format following the open-source code provided with the original papers; as such, they are slightly adjusted from the equations in the original papers.}. For each sample $\vx_i$, two augmentations are created by timestamp masking and random cropping, and then encoded as two representations $\vz_i^\prime$ and $\vz_i^{\dprime}$. TS2Vec and SoftCLT losses consider the same sum of similarities for a sample $i$ at a timestamp $t$, as shown in Equations (\ref{eq: sum_inst}) and (\ref{eq: sum_temp}). Equation (\ref{eq: sum_inst}) is used for instance-wise contrasting, which we denote by the subscript $_{\text{inst}}$; Equation (\ref{eq: sum_temp}) is used for time-wise contrasting, denoted by the subscript $_{\text{temp}}$.
\begin{equation}\label{eq: sum_inst}
    S_\text{inst}(i,t) = \sum_{j=1}^B\left(\exp(\vz^\prime_{i,t}\cdot \vz^{\dprime}_{j,t})+\exp(\vz^{\dprime}_{i,t}\cdot \vz^\prime_{j,t})\right)+\sum_{\substack{j=1\\j\neq i}}^B\left(\exp(\vz^\prime_{i,t}\cdot \vz^\prime_{j,t})+\exp(\vz^{\dprime}_{i,t}\cdot \vz^{\dprime}_{j,t})\right)
\end{equation}

\begin{equation}\label{eq: sum_temp}
    S_\text{temp}(i,t) = \sum_{s=1}^T\left(\exp(\vz^\prime_{i,t}\cdot \vz^{\dprime}_{i,s})+\exp(\vz^{\dprime}_{i,t}\cdot \vz^\prime_{i,s})\right)+\sum_{\substack{s=1\\s\neq t}}^T\left(\exp(\vz^\prime_{i,t}\cdot \vz^\prime_{i,s})+\exp(\vz^{\dprime}_{i,t}\cdot \vz^{\dprime}_{i,s})\right)
\end{equation}

Equation (\ref{eq: ts2vec_loss}) then shows the TS2Vec loss. We refer the readers to~\cite{Yue2022ts2vec} for more details about the hierarchical contrasting method.
\begin{equation}\label{eq: ts2vec_loss}
\begin{aligned}
    &\gL_\text{TS2Vec} = \frac{1}{NT}\sum_i\sum_t\left(\ell_{\substack{\text{inst}\\\text{TS2Vec}}}^{(i,t)}+\ell_{\substack{\text{temp}\\\text{TS2Vec}}}^{(i,t)}\right),\\
    \text{where } &
    \begin{cases}
        \displaystyle \ell_{\substack{\text{inst}\\\text{TS2Vec}}}^{(i,t)}=-\log\frac{\exp(\vz^\prime_{i,t}\cdot \vz^{\dprime}_{i,t})+\exp(\vz^{\dprime}_{i,t}\cdot \vz^{\prime}_{i,t})}{S_\text{inst}(i,t)}\\
        \displaystyle \ell_{\substack{\text{temp}\\\text{TS2Vec}}}^{(i,t)}=-\log\frac{\exp(\vz^\prime_{i,t}\cdot \vz^{\dprime}_{i,t})+\exp(\vz^{\dprime}_{i,t}\cdot \vz^{\prime}_{i,t})}{S_\text{temp}(i,t)}
    \end{cases}
\end{aligned}
\end{equation}

Similarity comparison in TS2Vec is between two different augmentations for the same sample. This is expanded by SoftCLT to also involve other samples in the same mini-batch. Varying instance-wise and time-wise weights are assigned to different comparison pairs as soft assignments, with Equations (\ref{eq: weight_inst}) and (\ref{eq: weight_temp}). This introduces four hyperparameters, i.e., $\tau_\text{inst}$, $\tau_\text{temp}$, $\alpha$, and $m$. We use EUC to compute $d(\vx_i,\vx_j)$ throughout this paper and set $\alpha=0.5$, both as recommended in the original paper; the other parameters need to be tuned for different datasets. Specifically, $m$ controls the sharpness of time hierarchical contrasting. TS2Vec uses $m=1$ (constant) and SoftCLT uses $m(k)=2^k$ (exponential), where $k$ is the depth of pooling layers when computing temporal loss. In this study, we add one more option $m(k)=k+1$ (linear), and will tune the best way for different datasets.
\begin{equation}\label{eq: weight_inst}
    w_\text{inst}(i,j) = \frac{2\alpha}{1+\exp(\tau_\text{inst}\cdot d(\vx_i,\vx_j)))}+
    \begin{cases}
        1-\alpha, &\text{if } i=j \\
        0, &\text{if } i\neq j
    \end{cases}
\end{equation}

\begin{equation}\label{eq: weight_temp}
    w_\text{temp}(t,s) = \frac{2}{1+\exp(\tau_\text{temp}\cdot m\cdot|t-s|)}
\end{equation}

Then Equation (\ref{eq: softclt_loss}) shows the SoftCLT loss, where we let $\lambda$ be 0.5 as recommended in the original paper. For a more detailed explanation and analysis, we refer the readers to~\cite{lee2024softclt}.
\begin{equation}\label{eq: softclt_loss}
\begin{aligned}
    &\gL_\text{SoftCLT} = \frac{1}{NT}\sum_i\sum_t\left(\lambda\ell_{\substack{\text{inst}\\\text{SoftCLT}}}^{(i,t)}+(1-\lambda)\ell_{\substack{\text{temp}\\\text{SoftCLT}}}^{(i,t)}\right), \\
    \text{where } &
    \begin{cases}
        \begin{aligned}
            \ell_{\substack{\text{inst}\\\text{SoftCLT}}}^{(i,t)} &= -\sum_{j=1}^B w_\text{inst}(i,j)\log\frac{\exp(\vz^\prime_{i,t}\cdot \vz^{\dprime}_{j,t})+\exp(\vz^{\dprime}_{i,t}\cdot \vz^{\prime}_{j,t})}{S_\text{inst}(i,t)} \\
            &\quad - \sum_{\substack{j=1\\j\neq i}}^B w_\text{inst}(i,j)\log\frac{\exp(\vz^\prime_{i,t}\cdot \vz^{\prime}_{j,t})+\exp(\vz^{\dprime}_{i,t}\cdot \vz^{\dprime}_{j,t})}{S_\text{inst}(i,t)}
        \end{aligned} \\
        \begin{aligned}
            \ell_{\substack{\text{temp}\\\text{SoftCLT}}}^{(i,t)} &= -\sum_{s=1}^T w_\text{temp}(t,s)\log\frac{\exp(\vz^\prime_{i,t}\cdot \vz^{\dprime}_{i,s})+\exp(\vz^{\dprime}_{i,t}\cdot \vz^{\prime}_{i,s})}{S_\text{temp}(i,t)} \\
            &\quad - \sum_{\substack{s=1\\s\neq t}}^T w_\text{temp}(t,s)\log\frac{\exp(\vz^\prime_{i,t}\cdot \vz^{\prime}_{i,s})+\exp(\vz^{\dprime}_{i,t}\cdot \vz^{\dprime}_{i,s})}{S_\text{temp}(i,t)}
        \end{aligned}        
    \end{cases}
\end{aligned}
\end{equation}

\subsection{Structure-preserving regularisers}
We use the topology-preserving loss proposed in~\cite{moor2020topoae} and the graph-geometry-preserving loss proposed in~\cite{lim2024ggae} as two structure-preserving regularisers, respectively focusing on the global and local structure of similarity relations. The global structure is preserved for instance-wise comparison, and the local structure is preserved for comparison across temporal or spatial features. In the following, we briefly describe the two losses, and the readers are referred to the original papers for more details.

Equation (\ref{eq: topo_loss}) presents the topology-preserving loss computed in each mini-batch. Here $\mA$ is a $B\times B$ EUC distance matrix between the samples in a batch, and is used to construct the Vietoris-Rips complex; $\boldsymbol{\pi}$ represents the persistence pairing indices of simplices that are considered topologically significant. The superscripts $X$ and $Z$ indicate original data space and latent space, respectively. 
\begin{equation}\label{eq: topo_loss}
    \gL_\text{Topo}=\frac{1}{2}\left\|\mA^X\left[\boldsymbol{\pi}^X\right]-\mA^Z\left[\boldsymbol{\pi}^X\right]\right\|^2+\frac{1}{2}\left\|\mA^Z\left[\boldsymbol{\pi}^Z\right]-\mA^X\left[\boldsymbol{\pi}^Z\right]\right\|^2
\end{equation}

The graph-geometry-preserving loss is also computed per mini-batch, as is shown in Equation (\ref{eq: ggeo_loss}). $\gL_\text{GGeo}$ measures geometry distortion, i.e., how much $f_{\boldsymbol{\theta}}$ deviates from being an isometry that preserves distances and angles. The geometry to be preserved of the original data manifold is implied by a similarity graph. To represent temporal and spatial characteristics, instead of using an instance as a node in the graph, we consider the nodes as timestamps or in a spatial dimension such as sensors or road users. Then the edges in the graph are defined by pairwise geodesic distances between nodes. 
\begin{equation}\label{eq: ggeo_loss}
    \gL_\text{GGeo}=\frac{1}{B}\sum_{i=1}^B \text{Tr}\left[\tilde{H}_i\left(L,\tilde{f}_\theta(\vx_i)\right)^2-2\tilde{H}_i\left(L,\tilde{f}_\theta(\vx_i)\right)\right],
\end{equation}
where $\tilde{H}_i$ represents an approximation of the Jacobian matrix of $f_{\boldsymbol{\theta}}$. Note that $\tilde{f}_\theta(\vx_i)$ as the latent representation of $\vx_i$ needs to maintain the node dimension. For example, if the nodes are considered as timestamps, $\tilde{f}_\theta(\vx_i)\in\sR^{T\times P}$; if the nodes are spatial objects, $\tilde{f}_\theta(\vx_i)\in\sR^{S\times P}$. With a similarity graph defined, then $L$ is the graph Laplacian that is approximated using a kernel matrix with width hyperparameter $h$, which requires tuning for different datasets.

\subsection{Stabilisation around the theoretical optimal values}\label{sec: optimal loss values}
This study considers $\gL_\text{TS2Vec}$ and $\gL_\text{SoftCLT}$ as $\gL_\text{CLT}$, and $\gL_\text{SP}$ can be $\gL_\text{Topo}$ or $\gL_\text{GGeo}$. Under this consideration, the optimal values for both $\gL_\text{CLT}$ and $\gL_\text{SP}$ are 0. For $\gL_\text{TS2Vec}$, a value of 0 is reached when $\vz_{i,t}^\prime$ and $\vz_{i,t}^{\dprime}$ are identical. Similarly, the optimal case of $\gL_\text{SoftCLT}$ is when the samples with soft assignments close to 1 are identical, while dissimilar samples have soft assignments close to 0. The topology-preserving loss $\gL_\text{Topo}$ is 0 when the topologically relevant distances remain the same in the latent space as in the original space, i.e., $\mA^X\left[\boldsymbol{\pi}^X\right]=\mA^Z\left[\boldsymbol{\pi}^X\right]$ and $\mA^X\left[\boldsymbol{\pi}^Z\right]=\mA^Z\left[\boldsymbol{\pi}^Z\right]$. Finally, $\gL_\text{GGeo}$ approximates the distortion measure of isometry and is ideally 0, although it may be negative as the approximation of $\tilde{H}_i$ is kernel-based depending on the hyperparameter $h$. 
\begin{figure}[htb]
    \centering
    \includegraphics[width=0.65\linewidth]{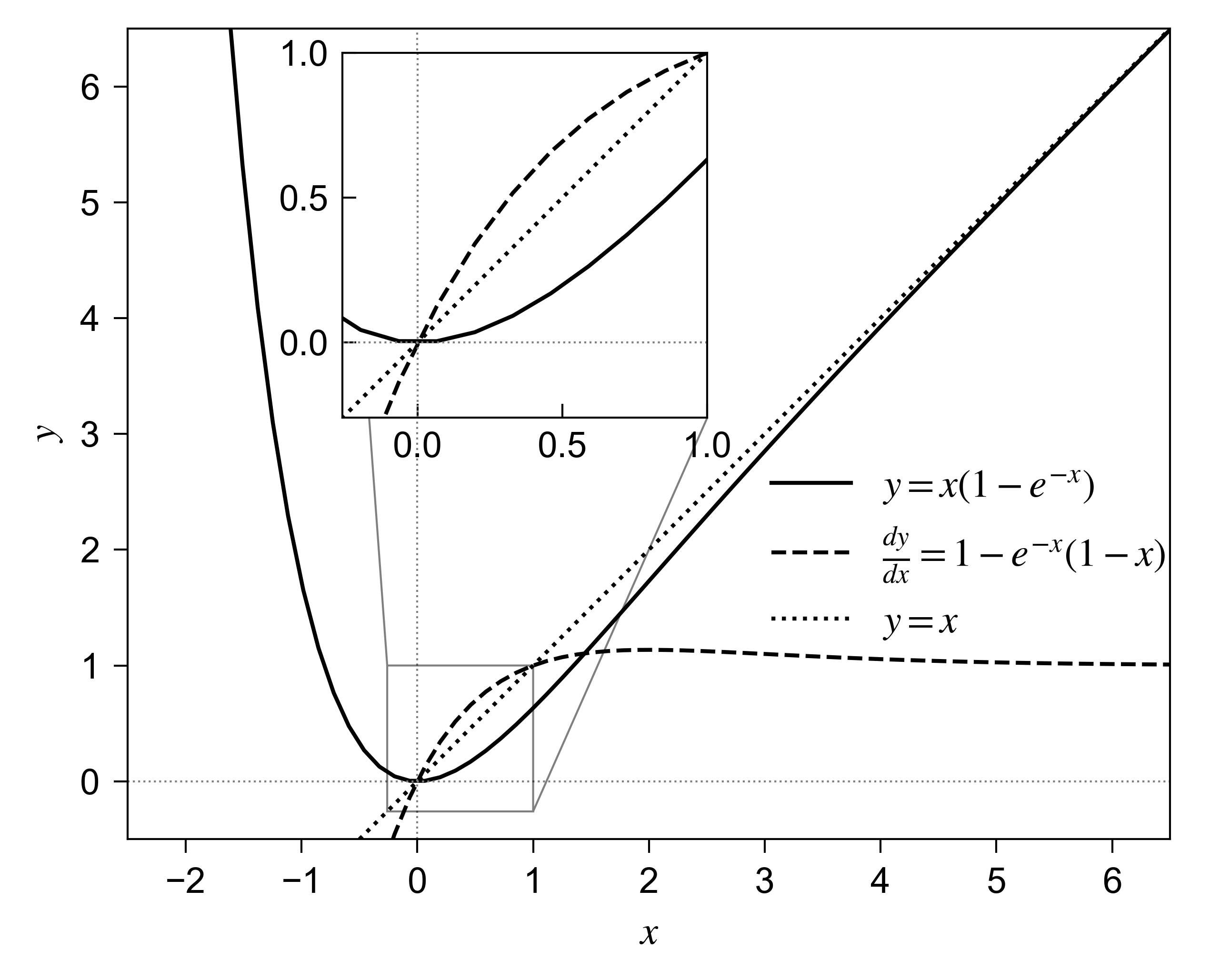}
    \caption{Illustration of the modification function.}
    \label{fig: modification}
\end{figure}

Section~\ref{sec: spclt loss} has briefly described the modification of $\gL_\text{CLT}$ and $\gL_\text{SP}$ by $y=x(1-\exp(-x))$ in Equation (\ref{eq: combined_loss}). Now we explain the effect of this function more with Figure~\ref{fig: modification}. The first objective of this modification is to penalise negative values of $\gL_\text{CLT}$ and $\gL_\text{SP}$. While $x>0$, $y$ decreases as $x$ decreases; but when $x<0$, $y$ rapidly increases as $x$ decreases. Once $\gL<0$, the modified term $\ell$ increases and the direction of gradient descent reverses. 

Another objective is to avoid large-step updates when $\gL$ is close to 0, so as not to miss its optimum. The modified $y$ approximates $x$ while $x$ is large, but has a slower decreasing rate when $x<1$. More specifically, the derivative of $x(1-\exp(-x))$ is $x^\prime(1-\exp(-x)(1-x))$, where $x^\prime$ denotes the derivative of $x$. While $x>1$, the multiplier in the parentheses is around 1 and $x^\prime$ is less  interfered with. While $x$ decreases from 1 to 0, the multiplier decreases from 1 to 0, so the derivative also decreases. As a result, this modification can stabilise training when either $\gL_\text{CLT}$ or $\gL_\text{SP}$ approaches its optimal value 0.

\subsection{Dynamic weighting mechanism to balance contrastive learning and structure preservation}
The training needs to balance between contrastive learning and structure preservation to avoid the neural network parameters being biased by either of the two objectives. However, the magnitudes of $\gL_\text{CLT}$ and $\gL_\text{SP}$ vary with different datasets and hyperparameter settings. This variation precludes fixed weights for the two modified losses. 

Inspired by~\cite{Kendall2018multitask}, we weigh $\ell_\text{CLT}$ and $\ell_\text{SP}$ by considering their uncertainties. We consider the loss values as deviations from their optimal values, and learn adaptive weights according to the deviations. Given the optimal value of 0, we assume a Gaussian distribution of $\ell$ with standard deviation $\sigma$, i.e., $p(\ell)=\gN(0, \sigma^2)$. Then we can maximise the log likelihood $\sum\log p(\ell)=\sum(-\log 2\pi-\log \sigma^2-\ell^2/\sigma^2)/2$ to learn $\sigma$. This is equivalent to minimising $\sum\left(\ell^2/2/\sigma^2+\log\sigma\right)$. When balancing between two losses $\ell_\text{CLT}$ and $\ell_\text{SP}$ that have deviations $\sigma_\text{CLT}$ and $\sigma_\text{SP}$,  respectively, we need to use Equation (\ref{eq: balanced_loss}).  
\begin{equation}\label{eq: balanced_loss}
    \argmax -\sum\log p(\ell_\text{CLT})p(\ell_\text{SP})\Leftrightarrow \argmin \sum\left(\frac{1}{2\sigma_\text{CLT}^2}\ell_\text{CLT}+\frac{1}{2\sigma_\text{SP}^2}\ell_\text{SP}+\log\sigma_\text{CLT}\sigma_\text{SP}\right)
\end{equation}

Replacing $\ell_\text{CLT}$ in Equation (\ref{eq: balanced_loss}) with $\gL_\text{CLT}\left(1-\exp(-\gL_\text{CLT})\right)$ and $\ell_\text{SP}$ with $\gL_\text{SP}\left(1-\exp(-\gL_\text{SP})\right)$, Equation (\ref{eq: combined_loss}) is eventually derived to be the complete loss. The training process trades-off between $\gL_\text{CLT}$ and $\gL_\text{SP}$, as well as between the weight regulariser $r_{\boldsymbol{\eta}}=\log\sigma_\text{CLT}\sigma_\text{SP}$ and the rest of Equation (\ref{eq: combined_loss}). When $\gL_\text{CLT}$ is small and $\gL_\text{SP}$ is large, $\sigma_\text{CLT}$ becomes small and $\sigma_\text{SP}$ becomes large, which then increases the weight for $\gL_\text{CLT}$ while reduces the weight for $\gL_\text{SP}$. The reverse occurs when $\gL_\text{CLT}$ is large and $\gL_\text{SP}$ is small. As the weighted sum of $\gL_\text{CLT}$ and $\gL_\text{SP}$ increases by larger weights, $\log\sigma_\text{CLT}\sigma_\text{SP}$ decreases and discourages the increase from being too much. Similarly, if the weighted sum decreases by smaller weights, $\log\sigma_\text{CLT}\sigma_\text{SP}$ also regularises the decrease.

\section{Experiments}\label{sec: exps}
We compare 6 losses for self-supervised representation learning (SSRL) of time series: TS2Vec, SoftCLT, Topo-TS2Vec, GGeo-TS2Vec, Topo-SoftCLT, and GGeo-SoftCLT. Among the losses, TS2Vec~\cite{Yue2022ts2vec} and SoftCLT~\cite{lee2024softclt} are baselines, and the others extend these two with a topology-preserving or a graph-geometry-preserving regulariser. The comparison is then evaluated by downstream task performances using these differently encoded representations. Consequently, the comparison and evaluation serve as an extensive ablation study focusing on the effects of structure-preserving regularisers. Our experiments are conducted with an NVIDIA A100 GPU with 80GB RAM and 5 Intel Xeon CPUs. For fair comparisons, we control the following conditions during experiments: random seed, the space and strategy for hyperparameter search, maximum training epochs, early stopping criteria, and samples used for evaluating local structure preservation.

\subsection{Baselines and datasets}
The evaluation of performance improvement is on 3 downstream tasks: multivariate time series classification, macroscopic traffic prediction, and microscopic traffic prediction. For every downstream task, we split training/(validation)/test sets following the baseline study and make sure the same data are used across models. Each experiment for a task has two stages, of which the first is SSRL and the second uses the encoded representations to perform classification/prediction. Only the split training set is used in the first stage, with 25\% separated as an internal validation set to schedule the learning rate for SSRL.

The classification task is on 28 datasets\footnote{The UEA archive collects 30 datasets in total. We omitted the two largest, InsectWingbeat and PenDigits, due to limited computation resources.} retrieved from the UEA archive~\cite{bagnall2018uea}. For each dataset, we set the representation dimension to 320 as used in the TS2Vec and SoftCLT studies, train 6 encoders with the 6 losses, and then classify the encoded representations with an RBF-kernel SVM. For traffic prediction, we use the dataset and model in~\cite{Li2024macro} for the macroscopic baseline, and those in~\cite{Li2024micro} for the microscopic baseline. The macroscopic traffic prediction uses 40 minutes (2-minute intervals) of historical data in 193 consecutive road segments to predict for all segments in the next 30 minutes. The microscopic traffic prediction forecasts the trajectory of an ego vehicle in 3 seconds, based on the history of up to 26 surrounding road users in the past 1 second (0.1-second intervals). Both traffic prediction baselines use encoder-decoder structures. We first pretrain the encoder with the 6 different losses for SSRL, and then fine-tune the complete model for prediction. The baseline trained from scratch is also compared.

To facilitate clearer analyses when presenting results, we divide the datasets included in the UEA archive into those with spatial features and those without. According to data descriptions in~\cite{bagnall2018uea}, the UEA datasets are grouped into 6 categories: human activity recognition, motion classification, ECG classification, EEG/MEG classification, audio spectra classification, and other problems. The human activity and motion categories, along with the PEMS-SF and LSST datasets that are categorised as other problems, contain spatial features. We thus consider these as spatial, and the remaining datasets as non-spatial. As a result, each division includes 14 datasets.

\subsection{Hyperparameters}
For each dataset, we perform a grid search to find the parameters that minimise $\gL_\text{CLT}$ after a certain number of iterations, where we set a constant learning rate of 0.001. Table \ref{tab: para_search_space} shows the search spaces of various hyperparameters, where bs is abbreviated for batch size and lr$_\eta$ is a separate learning rate for dynamic weights. When searching for best-suited parameters, we first set them as default values, and then follow the search strategy presented in Table \ref{tab: para_search_strategy}.
\begin{table}[htbp]
\begin{center}
\begin{minipage}{0.65\textwidth}
\renewcommand{\thempfootnote}{\alph{mpfootnote}}
\renewcommand*\footnoterule{}
\begin{center}
\caption{Hyperparameter search space.}
\label{tab: para_search_space}
\begin{tabular}{lll}
\toprule
 & Default & Search space \\ \midrule
bs & 8 & [8, 16, 32] \footnote{Maximum bs does not exceed train size.} \\
lr$_\eta$ & 0.05 & [0.01, 0.05] \\
$h$ & 1 & [0.25, 1, 9, 25, 49] \\
$\tau_\text{temp}$ & 0 & [0.5, 1, 1.5, 2, 2.5] \\
$m$ & constant & [constant, linear, exponential] \\
$\tau_\text{inst}$ & 0 & [1, 3, 5, 10, 20] \\ \bottomrule
\multicolumn{3}{l}{bs: batch size; lr$_\eta$: learning rate for dynamic weights.}
\end{tabular}
\end{center}
\end{minipage}
\end{center}
\end{table}

\begin{table}[htbp]
\begin{center}
\caption{Hyperparameter search strategy.}
\label{tab: para_search_strategy}
\begin{tabular}{lcccccc}
\toprule
Stage & bs & lr$_\eta$ & $h$ & $\tau_\text{temp}$ & $m$ & $\tau_\text{inst}$ \\ \midrule
TS2Vec & $\triangle$ &  &  &  &  &  \\
Topo-TS2Vec & $\square$ & $\triangle$ &  &  &  &  \\
GGeo-TS2Vec & $\square$ & $\triangle$ & $\triangle$ &  &  &  \\
SoftCLT Phase 1 & $\bigcirc$ &  &  & $\triangle$ & $\triangle$ & $\bigcirc$ \\
SofrCLT Phase 2 & $\triangle$ &  &  & $\square$ & $\square$ & $\triangle$ \\
Topo-SoftCLT & $\square$ & $\triangle$ &  & $\square$ & $\square$ & $\square$ \\
GGeo-SoftCLT & $\square$ & $\triangle$ & $\triangle$ & $\square$ & $\square$ & $\square$ \\ 
\bottomrule
\multicolumn{7}{l}{$\bigcirc$: default; $\square$: inherited; $\triangle$: tuned.}
\end{tabular}
\end{center}
\end{table}

The search spaces and strategy can result in up to 63 runs for one dataset. To save searching time, we adjust the number of iterations to be adequate to reflect the progress of loss reduction but limited to prevent overfitting, as our goal is to identify suitable parameters rather than fully train the models. The number of iterations is scaled according to the number of training samples, with larger datasets receiving more iterations. 

\subsection{Evaluation metrics}
Our performance evaluation uses both task-specific metrics and structure-preserving metrics. The former serves to validate performance improvements, while the latter serves to verify the effectiveness of preserving similarity structures. These metrics differ in whether a higher or lower value signifies better performance. To consistently indicate the best method, in the tables presented in the following sub-sections, the \textbf{\underline{best}} values are both bold and underlined; the \textbf{second-best} values are bold.

For evaluating the classification task, we use accuracy (Acc.) and the area under the precision-recall curve (AUPRC). To evaluate macroscopic traffic prediction, we use mean absolute error (MAE), root mean squared error (RMSE), the standard deviation of prediction errors (SDEP), and the explained variance by prediction (EVar). Dealing with microscopic traffic, we predict vehicle trajectories and assess the minimum final displacement error (min. FDE) as well as missing rates under radius thresholds of 0.5 m, 1 m, and 2 m (MR$_{0.5}$, MR$_{1}$, and MR$_{2}$). 

As for metrics to evaluate structure preservation, we adopt a combination of those used in~\cite{moor2020topoae} and~\cite{lim2024ggae}. More specifically, we consider 1) kNN, the proportion of shared k-nearest neighbours according to distance matrices in the latent space and in the original space; 2) continuity (Cont.), one minus the proportion of neighbours in the original space that are no longer neighbours in the latent space; 3) trustworthiness (Trust.), the counterpart of continuity, measuring the proportion of neighbours in the latent space but not in the original space; 4) MRRE, the averaged error in the relative ranks of sample distances between in the latent and original space; and 5) distance matrix RMSE (dRMSE), the root mean squared difference between sample distance matrices in the latent and original space. We calculate these metrics at two scales to evaluate global and local structure preservation. For global evaluation, our calculation is based on the EUC distances between samples; for local evaluation, it is based on the EUC distances between timestamps in a sample for at most 500 samples in the test set. 

\section{Results}\label{sec: results}
\subsection{Multivariate time series classification}
Table \ref{tab: uea_task} displays the classification performance on spatial and non-spatial UEA datasets. Next to the averaged accuracy, we also include the loss values on test sets to offer more information. More detailed results can be found in Tables \ref{tab: uea_spatial_details} and \ref{tab: uea_nonspatial_details} in \ref{sec: apdx_UEA_details}, where we present the classification accuracy with different representation learning losses for each dataset. Then we use Table \ref{tab: uea_improvement} to more specifically compare the relative improvements induced by adding a topology or graph-geometry preserving regulariser. The improvement is measured by the percentage of accuracy difference from the corresponding baseline performance.
\begin{table}[htb]
\begin{center}
\caption{UEA classification evaluation.}
\label{tab: uea_task}
\begin{tabular}{clcccc}
\toprule
Datasets & \multicolumn{1}{c}{Method} & Acc. (↑) & AUPRC (↑) & $\gL_\text{CLT}$ & $\gL_\text{SP}$ \\ \midrule
\multirow{6}{*}{\begin{tabular}[c]{@{}c@{}}With spatial\\ features (14)\end{tabular}} & TS2Vec & 0.848 & 0.872 & 2.943 &  \\
 & Topo-TS2Vec & 0.851 & 0.876 & 2.264 & 0.085 \\
 & GGeo-TS2Vec & 0.856 & 0.881 & 2.200 & 186.9 \\
 & SoftCLT & 0.852 & 0.876 & 7.943 &  \\
 & Topo-SoftCLT & \textbf{0.862} & \textbf{0.882} & 4.900 & 0.087 \\
 & GGeo-SoftCLT & \textbf{\underline{0.864}} & \textbf{\underline{0.883}} & 2.316 & 221.1 \\ \midrule
\multirow{6}{*}{\begin{tabular}[c]{@{}c@{}}Without spatial\\ features (14)\end{tabular}} & TS2Vec & 0.523 & 0.555 & 8.417 &  \\
 & Topo-TS2Vec & \textbf{\underline{0.553}} & \textbf{0.561} & 11.12 & 0.122 \\
 & GGeo-TS2Vec & 0.536 & \textbf{\underline{0.564}} & 15.58 & 957.0 \\
 & SoftCLT & 0.508 & 0.532 & 4.714 &  \\
 & Topo-SoftCLT & 0.496 & 0.534 & 7.328 & 0.124 \\
 & GGeo-SoftCLT & \textbf{0.537} & 0.549 & 10.09 & 144.7 \\ 
\bottomrule
\multicolumn{6}{l}{{\small Note: the \textbf{\underline{best}} values are both bold and underlined; the \textbf{second-best} values are bold.}}
\end{tabular}
\end{center}
\end{table}

\begin{table}[htb]
\begin{center}
\caption{Classification accuracy improved by Topo/GGeo regulariser. The comparisons are made with corresponding baseline performances.}
\label{tab: uea_improvement}
\begin{tabular}{ccccc}
\toprule
\multirow{2}{*}{Datasets} & \multicolumn{1}{c}{\multirow{2}{*}{Improvement by method}} & \multicolumn{3}{c}{Persentage in Acc. (\%)} \\ \cmidrule(l){3-5} 
 & \multicolumn{1}{c}{} & min. & mean & max. \\ \midrule
\multirow{4}{*}{\begin{tabular}[c]{@{}c@{}}With spatial\\ features (14)\end{tabular}} & Topo-TS2Vec & -4.403 & 0.800 & 16.54 \\
 & GGeo-TS2Vec & -3.783 & 1.143 & 10.44 \\
 & Topo-SoftCLT & -4.375 & 2.121 & 25.94 \\
 & GGeo-SoftCLT & -5.674 & 2.959 & 28.55 \\ \midrule
\multirow{4}{*}{\begin{tabular}[c]{@{}c@{}}Without spatial\\ features (14)\end{tabular}} & Topo-TS2Vec & -5.263 & 8.852 & 50.00 \\
 & GGeo-TS2Vec & -33.33 & 2.083 & 44.44 \\
 & Topo-SoftCLT & -33.33 & -0.815 & 50.00 \\
 & GGeo-SoftCLT & -20.83 & 18.49 & 166.7 \\ \bottomrule
\end{tabular}
\end{center}
\end{table}

Tables \ref{tab: uea_task} and \ref{tab: uea_improvement} clearly show that structure preservation improves classification accuracy, not only when time series data involve spatial features, but also when they do not. The relative improvements in Table \ref{tab: uea_improvement} are higher for non-spatial datasets than for spatial datasets, which is because the datasets without spatial features are more difficult to learn in the UEA archive. As is shown in Table \ref{tab: uea_task}, the loss of contrastive learning \textit{decreases} when a structure-preserving regulariser is added for spatial datasets, while \textit{increases} for non-spatial datasets. This implies that preserving similarity structure is well aligned with contrastive learning for spatial datasets, and can even enhance contrastive learning.

\begin{table}[htb]
\caption{Structure preservation evaluation over datasets with and without spatial features in the UEA archive. The standard deviations are computed across datasets.}
\label{tab: uea_dist_dens}
\begin{center}
\resizebox{\textwidth}{!}{%
\begin{tabular}{@{}clcccccccccc@{}}
\toprule
\multirow{2}{*}{Datasets} & \multicolumn{1}{c}{\multirow{2}{*}{Method}} & \multicolumn{5}{c}{Local mean between timestamps} & \multicolumn{5}{c}{Global mean between all samples} \\ \cmidrule(l){3-7}\cmidrule(l){8-12} 
 & \multicolumn{1}{c}{} & kNN (↑) & Trust. (↑) & Cont. (↑) & MRRE (↓) & dRMSE (↓) & kNN (↑) & Trust. (↑) & Cont. (↑) & MRRE (↓) & dRMSE (↓) \\ \midrule
\multirow{6}{*}{\begin{tabular}[c]{@{}c@{}}With\\ spatial\\ features\\ (14)\end{tabular}}
 & TS2Vec & 0.563 $\pm$ 0.149 & 0.875 $\pm$ 0.082 & 0.868 $\pm$ 0.073 & 0.117 $\pm$ 0.08 & 0.346 $\pm$ 0.099 & 0.419 $\pm$ 0.149 & 0.765 $\pm$ 0.110 & 0.784 $\pm$ 0.126 & 0.189 $\pm$ 0.124 & \textbf{0.150} $\pm$ 0.065 \\
 & Topo-TS2Vec & \textbf{0.569} $\pm$ 0.151 & 0.878 $\pm$ 0.082 & \textbf{0.873} $\pm$ 0.075 & \textbf{0.114} $\pm$ 0.081 & 0.344 $\pm$ 0.100 & 0.418 $\pm$ 0.150 & 0.764 $\pm$ 0.110 & 0.783 $\pm$ 0.125 & 0.190 $\pm$ 0.126 & 0.154 $\pm$ 0.069 \\
 & GGeo-TS2Vec & \textbf{0.569} $\pm$ 0.154 & \textbf{0.881} $\pm$ 0.08 & \textbf{0.873} $\pm$ 0.076 & \textbf{0.114} $\pm$ 0.079 & \textbf{0.341} $\pm$ 0.096 & 0.418 $\pm$ 0.153 & 0.762 $\pm$ 0.113 & 0.781 $\pm$ 0.129 & 0.190 $\pm$ 0.127 & 0.157 $\pm$ 0.075 \\
 & SoftCLT & 0.562 $\pm$ 0.156 & 0.875 $\pm$ 0.078 & 0.866 $\pm$ 0.074 & 0.117 $\pm$ 0.079 & 0.348 $\pm$ 0.096 & 0.420 $\pm$ 0.153 & 0.765 $\pm$ 0.112 & \textbf{0.788} $\pm$ 0.123 & \textbf{0.187} $\pm$ 0.125 & 0.171 $\pm$ 0.103 \\
 & Topo-SoftCLT & 0.564 $\pm$ 0.156 & 0.877 $\pm$ 0.077 & 0.869 $\pm$ 0.074 & 0.115 $\pm$ 0.077 & 0.344 $\pm$ 0.097 & \textbf{0.421} $\pm$ 0.157 & \textbf{0.767} $\pm$ 0.114 & 0.784 $\pm$ 0.126 & 0.188 $\pm$ 0.128 & 0.153 $\pm$ 0.068 \\
 & GGeo-SoftCLT & \textbf{\underline{0.571}} $\pm$ 0.150 & \textbf{\underline{0.883}} $\pm$ 0.073 & \textbf{\underline{0.875}} $\pm$ 0.067 & \textbf{\underline{0.111}} $\pm$ 0.076 & \textbf{\underline{0.337}} $\pm$ 0.091 & \textbf{\underline{0.425}} $\pm$ 0.149 & \textbf{\underline{0.768}} $\pm$ 0.110 & \textbf{\underline{0.790}} $\pm$ 0.121 & \textbf{\underline{0.185}} $\pm$ 0.125 & \textbf{\underline{0.149}} $\pm$ 0.065 \\
\midrule
\multirow{6}{*}{\begin{tabular}[c]{@{}c@{}}Without\\ spatial\\ features\\ (14)\end{tabular}}
 & TS2Vec & 0.423 $\pm$ 0.125 & \textbf{\underline{0.835}} $\pm$ 0.095 & \textbf{0.820} $\pm$ 0.105 & 0.150 $\pm$ 0.105 & \textbf{\underline{0.304}} $\pm$ 0.162 & \textbf{0.362} $\pm$ 0.175 & 0.767 $\pm$ 0.132 & 0.767 $\pm$ 0.136 & \textbf{\underline{0.252}} $\pm$ 0.151 & 0.197 $\pm$ 0.136 \\
 & Topo-TS2Vec & 0.424 $\pm$ 0.125 & 0.831 $\pm$ 0.100 & \textbf{0.820} $\pm$ 0.106 & 0.151 $\pm$ 0.105 & \textbf{0.308} $\pm$ 0.164 & 0.356 $\pm$ 0.176 & 0.767 $\pm$ 0.134 & 0.763 $\pm$ 0.139 & 0.254 $\pm$ 0.154 & \textbf{0.191} $\pm$ 0.120 \\
 & GGeo-TS2Vec & 0.420 $\pm$ 0.126 & 0.832 $\pm$ 0.094 & \textbf{0.820} $\pm$ 0.108 & 0.151 $\pm$ 0.105 & 0.310 $\pm$ 0.145 & \textbf{\underline{0.365}} $\pm$ 0.176 & \textbf{\underline{0.771}} $\pm$ 0.137 & \textbf{\underline{0.769}} $\pm$ 0.136 & 0.253 $\pm$ 0.157 & \textbf{\underline{0.189}} $\pm$ 0.125 \\
 & SoftCLT & \textbf{\underline{0.432}} $\pm$ 0.127 & \textbf{\underline{0.835}} $\pm$ 0.099 & \textbf{0.820} $\pm$ 0.106 & \textbf{0.148} $\pm$ 0.105 & 0.312 $\pm$ 0.162 & 0.354 $\pm$ 0.177 & 0.764 $\pm$ 0.134 & 0.763 $\pm$ 0.136 & \textbf{\underline{0.252}} $\pm$ 0.151 & 0.197 $\pm$ 0.131 \\
 & Topo-SoftCLT & 0.426 $\pm$ 0.119 & 0.834 $\pm$ 0.095 & 0.818 $\pm$ 0.104 & \textbf{0.148} $\pm$ 0.102 & 0.312 $\pm$ 0.163 & 0.361 $\pm$ 0.181 & \textbf{0.768} $\pm$ 0.131 & \textbf{0.768} $\pm$ 0.132 & 0.254 $\pm$ 0.152 & 0.205 $\pm$ 0.123 \\
 & GGeo-SoftCLT & \textbf{0.430} $\pm$ 0.122 & \textbf{\underline{0.835}} $\pm$ 0.095 & \textbf{\underline{0.822}} $\pm$ 0.101 & \textbf{\underline{0.147}} $\pm$ 0.101 & 0.315 $\pm$ 0.151 & 0.355 $\pm$ 0.174 & 0.762 $\pm$ 0.134 & 0.761 $\pm$ 0.136 & 0.257 $\pm$ 0.153 & 0.203 $\pm$ 0.132 \\
\bottomrule
\multicolumn{12}{l}{Note: the \textbf{\underline{best}} values are both bold and underlined; the \textbf{second-best} values are bold.}
\end{tabular}}
\end{center}
\end{table}

The assessment of similarity preservation is presented in Table \ref{tab: uea_dist_dens} at both local and global scales. Consistent with the task-specific evaluation, Table \ref{tab: uea_dist_dens} shows that structure-preserving regularisation preserves more complete information on similarity relations. The improvements are generally more significant on datasets with spatial features, which makes it more evident that our proposed preservation suits spatial time series data better. Although the comparisons in Tables \ref{tab: uea_task}$\sim$\ref{tab: uea_dist_dens} indicate more notable improvements by preserving graph geometry than preserving topology, we have to note that this does not demonstrate the universal superiority of one over the others. Different datasets have different characteristics that benefit from preserving global or local structure, and domain knowledge is necessary to determine which could be more effective. We will discuss this more in Section \ref{sec: discussion}.

\subsection{Macroscopic and microscopic traffic prediction}
\begin{figure}
    \centering
    \includegraphics[width=\linewidth]{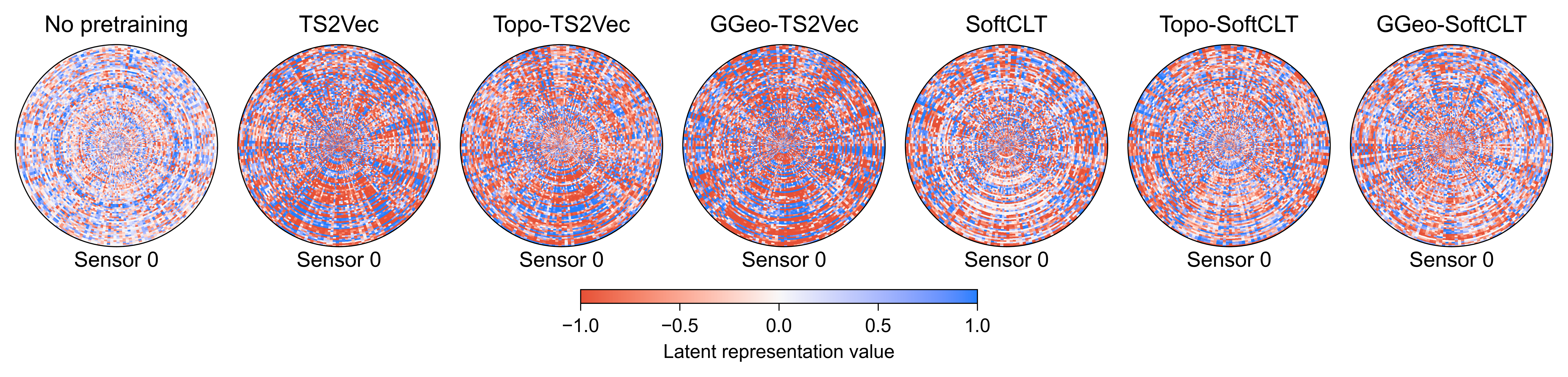}
    \caption{Encoded representations after training with different losses on the test set of the Macroscopic traffic prediction task.}
    \label{fig: latent_macro}
\end{figure}

In Table \ref{tab: traffic_task}, we present the performance evaluation for both macroscopic and microscopic traffic prediction. This table shows consistent improvements by pretraining encoders with our methods. Notably, single contrastive learning (i.e., TS2Vec and SoftCLT) does not necessarily improve downstream prediction, whereas it does when used together with preserving certain similarity structures. Given that our comparisons are conducted through controlling random conditions, this result effectively shows the necessity of preserving structure when learning traffic interaction representations. In addition, we plot polar heatmaps in Figure \ref{fig: latent_macro} to visualise the encoded latent representations for the sensors in macroscopic traffic prediction. These sensors are deployed along a ring road, thus adjacent sensors are expected to have similar states and representation patterns. The figure intuitively shows better preserved spatial-temporal relations by contrastive learning and structure preservation.
\begin{table}[htb]
\caption{Macroscopic and microscopic traffic prediction performance evaluation. Metrics are reported as mean $\pm$ standard deviation over 10 stratified folds of the test set.}
\label{tab: traffic_task}
\begin{center}
\resizebox{\textwidth}{!}{%
\begin{tabular}{@{}lcccccccc@{}}
\toprule
\multicolumn{1}{c}{\multirow{3}{*}{Method}} & \multicolumn{4}{c}{Macroscopic Traffic} & \multicolumn{4}{c}{Microscopic Traffic} \\ \cmidrule(l){2-5}\cmidrule(l){6-9} 
\multicolumn{1}{c}{} & \begin{tabular}[c]{@{}c@{}}MAE (↓)\\ (km/h)\end{tabular} & \begin{tabular}[c]{@{}c@{}}RMSE (↓)\\ (km/h)\end{tabular} & \begin{tabular}[c]{@{}c@{}}SDEP (↓)\\ (km/h)\end{tabular} & \begin{tabular}[c]{@{}c@{}}EVar (↑)\\ (\%)\end{tabular}  & \begin{tabular}[c]{@{}c@{}}min. FDE (↓)\\ (m)\end{tabular} & \begin{tabular}[c]{@{}c@{}}MR$_{0.5}$ (↓)\\ (\%)\end{tabular} & \begin{tabular}[c]{@{}c@{}}MR$_1$ (↓)\\ (\%)\end{tabular} & \begin{tabular}[c]{@{}c@{}}MR$_2$ (↓)\\ (\%)\end{tabular} \\ \midrule
No pretraining & \textbf{2.850} $\pm$ 0.044 & 5.911 $\pm$ 0.100 & 5.909 $\pm$ 0.100 & 84.784 $\pm$ 0.314 & 0.640 $\pm$ 0.013 & 59.253 $\pm$ 1.099 & 12.161 $\pm$ 0.929 & 0.744 $\pm$ 0.256 \\
TS2Vec & 2.878 $\pm$ 0.046 & 5.981 $\pm$ 0.111 & 5.981 $\pm$ 0.111 & 84.412 $\pm$ 0.337 & \textbf{0.636} $\pm$ 0.008 & 59.453 $\pm$ 0.866 & 11.899 $\pm$ 0.721 & \textbf{\underline{0.558}} $\pm$ 0.17 \\
Topo-TS2Vec & 2.862 $\pm$ 0.043 & 5.914 $\pm$ 0.099 & 5.907 $\pm$ 0.099 & 84.793 $\pm$ 0.306 & \textbf{\underline{0.634}} $\pm$ 0.008 & \textbf{\underline{58.144}} $\pm$ 1.014 & \textbf{11.761} $\pm$ 0.809 & 0.737 $\pm$ 0.252 \\
GGeo-TS2Vec & 2.887 $\pm$ 0.045 & 5.980 $\pm$ 0.107 & 5.977 $\pm$ 0.107 & 84.433 $\pm$ 0.326 & \textbf{0.636} $\pm$ 0.010 & \textbf{58.289} $\pm$ 1.007 & \textbf{\underline{11.747}} $\pm$ 0.797 & 0.737 $\pm$ 0.239 \\
SoftCLT & 2.856 $\pm$ 0.043 & 5.937 $\pm$ 0.105 & 5.931 $\pm$ 0.104 & 84.670 $\pm$ 0.348 & 0.641 $\pm$ 0.018 & 59.501 $\pm$ 1.742 & 11.940 $\pm$ 0.998 & 0.785 $\pm$ 0.278 \\
Topo-SoftCLT & \textbf{2.850} $\pm$ 0.045 & \textbf{5.881} $\pm$ 0.111 & \textbf{5.880} $\pm$ 0.111 & \textbf{84.934} $\pm$ 0.327 & 0.640 $\pm$ 0.012 & 58.626 $\pm$ 0.995 & 12.043 $\pm$ 0.735 & 0.820 $\pm$ 0.249 \\
GGeo-SoftCLT & \textbf{\underline{2.834}} $\pm$ 0.046 & \textbf{\underline{5.878}} $\pm$ 0.116 & \textbf{\underline{5.877}} $\pm$ 0.116 & \textbf{\underline{84.952}} $\pm$ 0.344 & 0.652 $\pm$ 0.013 & 60.638 $\pm$ 0.596 & 13.249 $\pm$ 0.681 & \textbf{0.723} $\pm$ 0.231 \\
Best improvement & 0.568 & 0.563 & 0.552 & 0.198 & 0.929 & 1.872 & 3.401 & 24.998 \\
\bottomrule
\multicolumn{9}{l}{Note: the \textbf{\underline{best}} values are both bold and underlined; the \textbf{second-best} values are bold.}
\end{tabular}}
\end{center}
\end{table}

Table \ref{tab: traffic_dist_dens} then displays the corresponding evaluation on similarity structure preservation, which is obtained by assessing the encoders after fine-tuning for traffic prediction. The results show that the better-performing methods in macro-traffic prediction (i.e., GGeo-SoftCLT, Topo-SoftCLT) and micro-traffic prediction (i.e., Topo-TS2Vec, GGeo-TS2Vec) preserve more similarity structures at both global and local scales. In general, the metric values are close for the same task across methods; however, when a method has a significant advantage over the others, it indicates superior performance. Examples for macroscopic traffic prediction are Topo-SoftCLT in local Cont. and GGeo-SoftCLT in global dRMSE; for microscopic traffic prediction are Topo-TS2Vec in local Trust. and GGeo-TS2Vec in global Cont.
\begin{table}[htb]
\caption{Structure preservation evaluation of encoders after the fine-tuning in traffic prediction tasks.}
\label{tab: traffic_dist_dens}
\begin{center}
\resizebox{\textwidth}{!}{%
\begin{tabular}{@{}lcccccccccc@{}}
\toprule
\multicolumn{1}{c}{\multirow{2}{*}{Method}} & \multicolumn{5}{c}{Macroscopic Traffic} & \multicolumn{5}{c}{Microscopic Traffic} \\ \cmidrule(l){2-6}\cmidrule(l){7-11} 
\multicolumn{1}{c}{} & kNN (↑) & Trust. (↑) & Cont. (↑) & MRRE (↓) & dRMSE (↓) & kNN (↑) & Trust. (↑) & Cont. (↑) & MRRE (↓) & dRMSE (↓) \\ \midrule
\multicolumn{11}{c}{Local mean between timestamps for at most 500 samples} \\ \midrule
No pretraining & 0.125 & 0.524 & \textbf{\underline{0.526}} & 0.496 & \textbf{\underline{0.224}} & 0.373 & 0.742 & 0.552 & 0.426 & \textbf{\underline{0.478}} \\
TS2Vec & 0.125 & 0.523 & 0.522 & 0.501 & 0.251 & \textbf{\underline{0.398}} & \textbf{\underline{0.761}} & \textbf{\underline{0.592}} & \textbf{\underline{0.393}} & 0.496 \\
Topo-TS2Vec & \textbf{0.128} & \textbf{0.533} & 0.522 & \textbf{\underline{0.491}} & \textbf{0.242} & 0.397 & 0.754 & \textbf{0.590} & 0.399 & 0.506 \\
GGeo-TS2Vec & 0.126 & 0.529 & \textbf{0.524} & 0.496 & 0.249 & 0.397 & \textbf{0.756} & 0.589 & \textbf{0.396} & 0.508 \\
SoftCLT & 0.127 & 0.526 & 0.523 & 0.500 & 0.246 & 0.378 & 0.746 & 0.552 & 0.427 & \textbf{\underline{0.478}} \\
Topo-SoftCLT & \textbf{\underline{0.129}} & \textbf{\underline{0.536}} & \textbf{0.524} & \textbf{0.492} & 0.250 & \textbf{\underline{0.398}} & 0.755 & 0.588 & 0.405 & 0.480 \\
GGeo-SoftCLT & 0.127 & 0.527 & 0.523 & 0.498 & 0.261 & 0.397 & 0.751 & 0.589 & 0.405 & 0.485 \\ \midrule
\multicolumn{11}{c}{Global mean between all samples} \\ \midrule
No pretraining & \textbf{\underline{0.316}} & \textbf{\underline{0.949}} & \textbf{\underline{0.969}} & \textbf{\underline{0.031}} & \textbf{0.364} & 0.218 & 0.937 & 0.920 & 0.049 & 0.141 \\
TS2Vec & 0.264 & 0.940 & 0.957 & 0.039 & 0.377 & 0.232 & 0.953 & 0.920 & \textbf{0.044} & \textbf{0.139} \\
Topo-TS2Vec & 0.276 & 0.942 & 0.963 & 0.036 & 0.379 & \textbf{0.233} & \textbf{0.958} & 0.917 & 0.045 & \textbf{\underline{0.138}} \\
GGeo-TS2Vec & 0.263 & 0.940 & 0.959 & 0.039 & 0.400 & 0.231 & \textbf{\underline{0.959}} & \textbf{0.923} & \textbf{\underline{0.041}} & 0.140 \\
SoftCLT & \textbf{0.299} & \textbf{0.943} & \textbf{0.966} & \textbf{0.035} & 0.391 & 0.224 & 0.924 & 0.916 & 0.055 & 0.148 \\
Topo-SoftCLT & 0.288 & 0.940 & 0.965 & 0.036 & 0.371 & 0.215 & 0.909 & 0.901 & 0.065 & 0.150 \\
GGeo-SoftCLT & 0.287 & 0.939 & 0.964 & 0.037 & \textbf{\underline{0.359}} & \textbf{\underline{0.240}} & 0.935 & \textbf{\underline{0.926}} & 0.046 & 0.146 \\ \bottomrule
\multicolumn{11}{l}{Note: the \textbf{\underline{best}} values are both bold and underlined; the \textbf{second-best} values are bold.}
\end{tabular}}
\end{center}
\end{table}

Notably, in macroscopic traffic prediction, fine-tuning from scratch maintains the greatest global similarities. This implies that the specific model architecture might allow for learning similarity structure without pretraining. This is not crystal clear with the final evaluation only. In the next sub-section, we will add different model architectures for the macro-traffic prediction task, and visualise the fine-tuning progress to further understand the contribution of structure preservation to downstream task performance.

\subsection{Training efficiency}
Incorporating structure-preserving regularisation increases computational complexity, and consequently, training time. The magnitude of this increase depends on the data and model that are applied on. With Table \ref{tab: training_time}, we quantify the additional time required for structure preservation and evaluate its impact across diverse model architectures. In prior experiments, we used Convolutional Neural Network (CNN) encoders for the classification task on UEA datasets, Dynamic Graph Convolution Network (DGCN,~\cite{Li2021dgcn}) encoder for macroscopic traffic prediction, and Hierarchical Graph Neural Network (HGNN, based on VectorNet~\cite{Gao2020,Gu2021}) encoder for microscopic traffic prediction. To obtain a more comprehensive evaluation, we include two more Recurrent Neural Network (RNN) models for macroscopic traffic prediction: Long Short-Term Memory (LSTM) and Gated Recurrent Unit (GRU) encoders, paired with simple linear decoders.
\begin{table}[htb]
\centering
\caption{Training time per epoch in the stage of self-supervised representation learning.}
\label{tab: training_time}
\begin{center}
\resizebox{\textwidth}{!}{%
\begin{tabular}{@{}ccccccccc@{}}
\toprule
Task/data & Encoder & Base (sec/epoch) & TS2Vec & Topo-TS2Vec & GGeo-TS2Vec & SoftCLT & Topo-SoftCLT & GGeo-SoftCLT \\ \midrule
\begin{tabular}[c]{@{}c@{}}Avg. UEA$^a$\end{tabular} & CNN & 11.94 & 1.00$\times$ & 1.46$\times$ & 2.35$\times$ & 1.00$\times$ & 1.46$\times$ & 2.36$\times$ \\
\begin{tabular}[c]{@{}c@{}}MicroTraffic\end{tabular} & HGNN & 122.93 & 1.00$\times$ & 1.45$\times$ & 1.16$\times$ & 1.23$\times$ & 1.69$\times$ & 1.42$\times$ \\
\multirow{3}{*}{\begin{tabular}[c]{@{}c@{}}MacroTraffic\end{tabular}} & DGCN & 74.78 & 1.00$\times$ & 1.29$\times$ & 1.09$\times$ & 1.02$\times$ & 1.40$\times$ & 1.37$\times$ \\
 & LSTM & 18.04 & 1.00$\times$ & 1.49$\times$ & 1.12$\times$ & 1.09$\times$ & 1.57$\times$ & 1.22$\times$ \\
 & GRU & 16.49 & 1.00$\times$ & 1.54$\times$ & 1.14$\times$ & 1.10$\times$ & 1.61$\times$ & 1.24$\times$ \\ \bottomrule
 \multicolumn{9}{l}{$^a$ Detailed results are referred to Tables \ref{tab: uea_spatial_details_time} and \ref{tab: uea_nonspatial_details_time} in \ref{sec: apdx_UEA_details}.}
\end{tabular}}
\end{center}
\vspace{-15pt}
\end{table}

Table \ref{tab: training_time} shows that preserving structure increases training time by less than 50\% in most cases, and suits DGCN particularly well with the least additional time. However, when time sequences are very long, the computation of graph-geometry preserving loss becomes intense. For example, the time series length is 1,197 in the Cricket dataset and results in a pretraining time of 2.86 times the base; likewise, the EthanolConcentration dataset has a length of 1,751 and a pretraining time of 4.09 times the base, and the EigenWorms dataset uses 7.59 times of the base pretraining time with a time series length of 17,984.

\begin{figure}[htbp]
\begin{center}
\includegraphics[width=\textwidth]{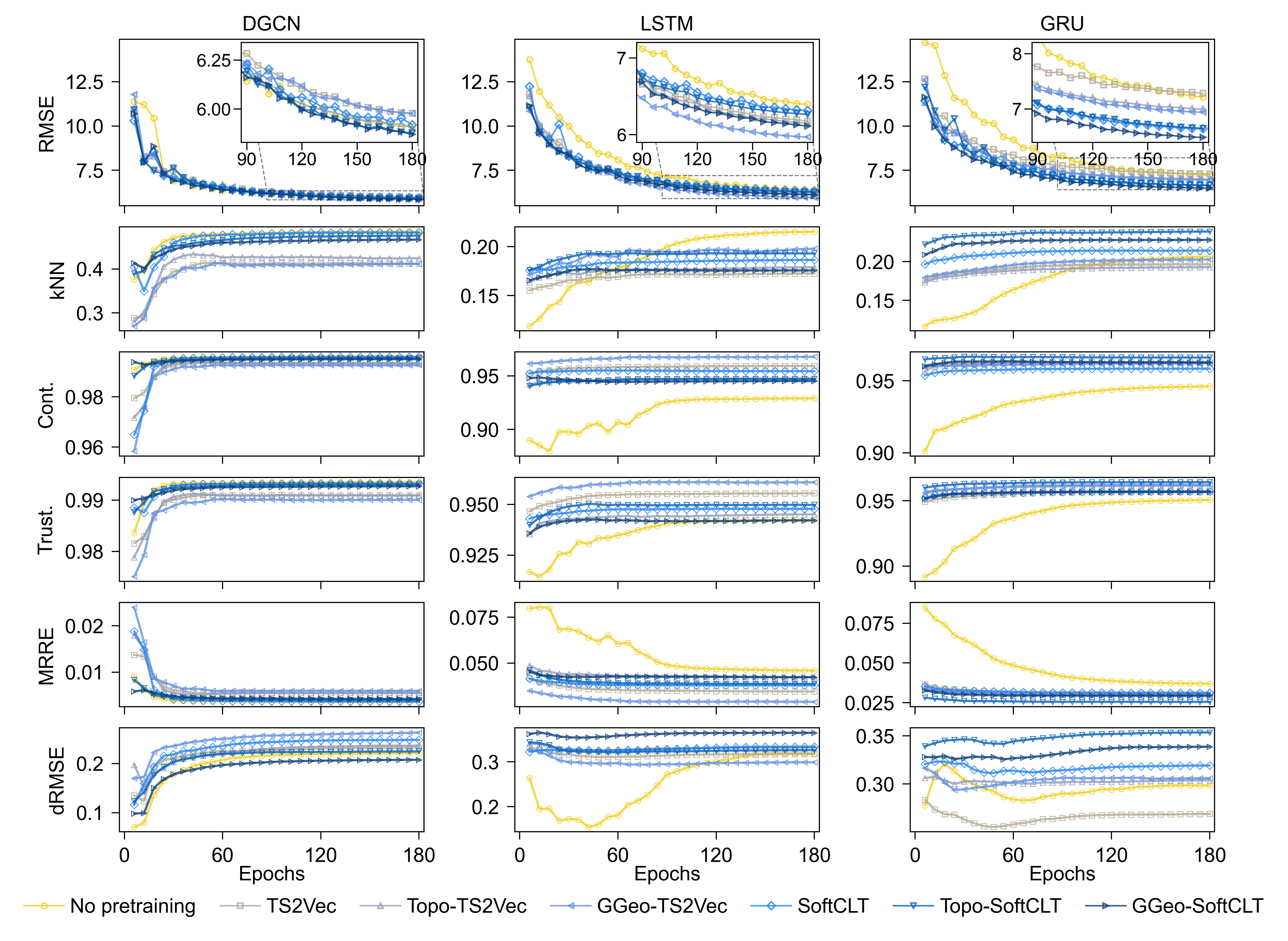}
\end{center}
\caption{Fine-tuning progress of models trained from scratch and pretrained with different losses in macroscopic traffic prediction. Values of the final performance are referred to Table \ref{tab: traffic_task} for DGCN, to Tables \ref{tab: lstm_gru_prediction} and \ref{tab: lstm_gru_structure} in \ref{sec: apdx_lstm_gru_details} for LSTM and GRU.}
\label{fig: macro_traffic_progress}
\end{figure}

In more detail, we evaluate the fine-tuning efficiency in macroscopic traffic prediction to further investigate the contribution of structure preservation. Figure \ref{fig: macro_traffic_progress} shows the convergence process of different models with and without pretraining, where RMSE is used to evaluate prediction performance and the other metrics indicate the preservation of global similarity relations.

For all models of DGCN, LSTM, and GRU, structure preservation consistently enhances prediction performance compared to training from scratch (No pretraining). The enhancement is significant when using LSTM and GRU, achieving 6.68\% and 10.14\% improvement in RMSE, respectively. Meanwhile, the progress of structure preservation is stable when the encoder is pretrained, and maintains the advantage over no pretraining throughout the fine-tuning process. In contrast, for DGCN, which is a more sophisticated model tailored for the task, training from scratch is already very effective and pretraining brings relatively minor improvement. This implies that certain model architectures are better suited to a specific task than others, and the preservation of similarity structure in the latent representations may be a good indicator for model selection.

\section{Discussion}\label{sec: discussion}
The theoretical foundation and experimental results presented in this paper not only demonstrate evident improvements in downstream task performance but also reveal a critical bridge between contrastive learning and similarity structure preservation. In this section, we discuss these findings to guide method selection, interpret the observed performance improvements, and remind potential failure modes.

\subsection{Method selection}
\begin{figure}[htb]
    \centering
    \includegraphics[width=4.in]{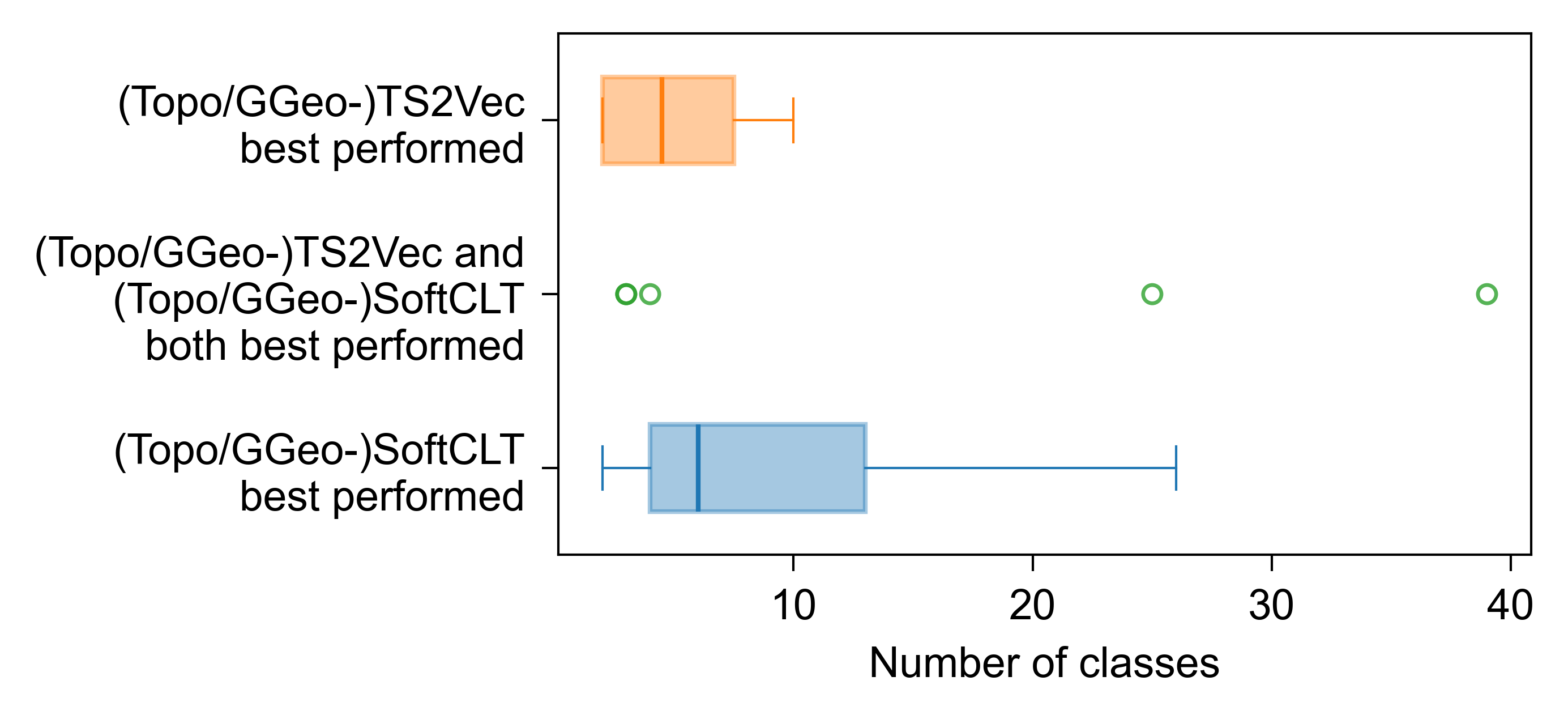}
    \caption{Box plots of the number of classes in the UEA datasets for which the best classification accuracy is achieved after contrastive learning based on TS2Vec or SoftCLT.}
    \label{fig: uea_num_classes}
\end{figure}

A key consideration when applying our method lies in selecting an appropriate loss function. This involves two layers of choices: TS2Vec versus SoftCLT, and topology-preserving (Topo-) versus graph-geometry-preserving (GGeo-) regularisation. For the first choice, TS2Vec is generally more suitable for classification tasks with fewer classes as TS2Vec only compares the similarity between two different augmentations of the same sample. In contrast, SoftCLT incorporates all samples in a mini-batch by assigning soft labels based on similarity. This performs a more detailed similarity comparison and thus is advantageous for tasks with a larger number of classes or for regression. As shown in Figure \ref{fig: uea_num_classes}, in the UEA archive, datasets for which (Topo/GGeo-)TS2Vec achieves the best accuracy tend to have fewer classes than those with the best performance achieved by (Topo/GGeo-)SoftCLT. In essence, SoftCLT implicitly embeds the similarity structure through soft labels.

Then the second layer of choice hinges on the scale of structural relevance to the downstream task. The topology-preserving regulariser is designed to maintain the global similarity structure between data samples, and the graph-geometry-preserving regulariser focuses on locally preserving temporal or spatial similarity structures within samples. Therefore, Topo-regularisation is especially beneficial when the downstream task relies on inter-sample relations; whereas GGeo-regularisation is particularly useful in tasks where subtle intra-sample variations are critical, such as traffic prediction.

Our hypothesis for the performance improvements observed across tasks is that explicit structure-preserving regularisation enforces the encoded latent space to reflect the original distribution of data. This aligns with the theoretical view of neural networks as models that approximate the conditional distributions of outputs on inputs. Contrastive learning typically relies on predefined positive and negative augmentations, which may disrupt the original patterns underlying data and introduce biases. By anchoring the latent space to the original data manifold, structure preservation can effectively mitigate the potential biases by reducing the dependency on augmentations. 

\subsection{Failure modes}

While the proposed method demonstrates consistent improvements across a range of tasks, several limitations and potential failure modes need to be acknowledged. These limitations are instructive in identifying the contexts where the method is most effective and where caution is warranted.

One source of difficulty arises from the sensitivity to data scale and complexity. The additional cost by structure-preserving regularisation is acceptable in many cases, but the overhead becomes prohibitive when time series are extremely long . For example, datasets with thousands of time steps result in a large similarity graph that inflates the computation memory and time of the GGeo-loss. Meanwhile, for models that are already well aligned with data, such as the DGCN baseline in macroscopic traffic prediction, pretraining with our method may provide only marginal benefits. This suggests that the utility of structure preservation is most pronounced when the data involve complex similarity relations that base models cannot easily capture.

\begin{table}[htb]
\caption{Comparison of dynamic and fixed weighting performance with the UEA archive. By fixed weighting we let $\sigma_\text{CLT}=1$ and $\sigma_\text{SP}=1$, so that the weights for both contrastive learning and structure preservation are 0.5.}
\label{tab: uea_weighting}
\begin{center}
\resizebox{\textwidth}{!}{%
\begin{tabular}{@{}llcccccccc@{}}
\toprule
\multicolumn{1}{c}{Dataset} & \multicolumn{1}{c}{Method} & Acc. (↑) & kNN (↑) & Trust. (↑)        & Cont. (↑) & MRRE (↓) & dRMSE (↓) & $\gL_\text{CLT}$ & $\gL_\text{SP}$ \\
\midrule
\multirow{2}{*}{ERing} & GGeo-SoftCLT (dynamic weighting) & 0.871 $\pm$ 0.011 & 0.776 $\pm$ 0.004 & 0.954 $\pm$ 0.002 & 0.940 $\pm$ 0.003  & 0.047 $\pm$ 0.001 & 0.209 $\pm$ 0.007 & 0.652 $\pm$ 0.023  & 106.972 $\pm$ 15.521    \\
& GGeo-SoftCLT (fixed weighting)   & 0.876 $\pm$ 0.013 & 0.767 $\pm$ 0.003 & 0.947 $\pm$ 0.002 & 0.933 $\pm$ 0.002 & 0.05 $\pm$ 0.001  & 0.226 $\pm$ 0.004 & 0.602 $\pm$ 0.015  & 149.115 $\pm$ 15.738    \\ \midrule
\multirow{2}{*}{Libras} & GGeo-TS2Vec (dynamic weighting)  & 0.863 $\pm$ 0.010  & 0.757 $\pm$ 0.002 & 0.945 $\pm$ 0.001 & 0.915 $\pm$ 0.001 & 0.091 $\pm$ 0.001 & 0.281 $\pm$ 0.004 & 4.724 $\pm$ 0.038  & 387.867 $\pm$ 47.765    \\
& GGeo-TS2Vec (fixed weighting)    & 0.864 $\pm$ 0.016 & 0.754 $\pm$ 0.002 & 0.943 $\pm$ 0.001 & 0.915 $\pm$ 0.001 & 0.091 $\pm$ 0.001 & 0.275 $\pm$ 0.004 & 4.818 $\pm$ 0.021  & 519.156 $\pm$ 33.440     \\ \midrule
\multirow{2}{*}{NATOPS} & GGeo-SoftCLT (dynamic weighting) & 0.921 $\pm$ 0.009 & 0.640 $\pm$ 0.004  & 0.903 $\pm$ 0.004 & 0.859 $\pm$ 0.004 & 0.108 $\pm$ 0.002 & 0.296 $\pm$ 0.004 & 0.450 $\pm$ 0.007   & 712.646 $\pm$ 39.097    \\
& GGeo-SoftCLT (fixed weighting)   & 0.920 $\pm$ 0.011  & 0.639 $\pm$ 0.003 & 0.902 $\pm$ 0.003 & 0.855 $\pm$ 0.003 & 0.109 $\pm$ 0.002 & 0.295 $\pm$ 0.004 & 0.447 $\pm$ 0.006  & 811.255 $\pm$ 55.068    \\ \midrule
\multirow{2}{*}{StandWalkJump} & GGeo-TS2Vec (dynamic weighting)  & 0.273 $\pm$ 0.049 & 0.389 $\pm$ 0.007 & 0.947 $\pm$ 0.003 & 0.905 $\pm$ 0.004 & 0.058 $\pm$ 0.002 & 0.226 $\pm$ 0.032 & 12.592 $\pm$ 2.747 & 3826.480 $\pm$ 2231.619  \\
& GGeo-TS2Vec (fixed weighting)    & 0.300 $\pm$ 0.057   & 0.386 $\pm$ 0.006 & 0.945 $\pm$ 0.003 & 0.904 $\pm$ 0.004 & 0.059 $\pm$ 0.002 & 0.239 $\pm$ 0.034 & 12.821 $\pm$ 2.847 & 4173.859 $\pm$ 2471.534 \\ \midrule
\multirow{2}{*}{UWaveGestureLibrary} & GGeo-TS2Vec (dynamic weighting)  & 0.855 $\pm$ 0.011 & 0.759 $\pm$ 0.009 & 0.950 $\pm$ 0.005  & 0.956 $\pm$ 0.006 & 0.03 $\pm$ 0.002  & 0.308 $\pm$ 0.007 & 1.098 $\pm$ 0.170   & 501.284 $\pm$ 47.744    \\
& GGeo-TS2Vec (fixed weighting)    & 0.853 $\pm$ 0.018 & 0.755 $\pm$ 0.009 & 0.948 $\pm$ 0.006 & 0.954 $\pm$ 0.006 & 0.031 $\pm$ 0.002 & 0.306 $\pm$ 0.007 & 1.102 $\pm$ 0.174  & 604.046 $\pm$ 67.679    \\
\bottomrule
\end{tabular}}
\end{center}
\end{table}

Another important consideration concerns the benefits by dynamic weighting compared to fixed weights. The dynamic weighting mechanism is introduced to adaptively balance the loss magnitudes of contrastive learning and structure preservation. This works particularly well when the loss components have a large magnitude gap, which biases parameter updates. However, dynamic weighting may not outperform fixed weighting when the magnitudes of loss components are balanced in nature. In such circumstances, the adaptive mechanism can oscillate excessively, delaying convergence or amplifying noise in the latent space. Table \ref{tab: uea_weighting} compares model performance on 5 datasets in the UEA archive. The datasets are selected because of relatively large magnitude gaps between contrastive learning loss and structure preservation loss. Dynamic and fixed weighting yield comparable classification accuracies, but the dynamic mechanism consistently improves structure preservation metrics and achieve lower loss.

Finally, dependency on hyperparameter tuning presents a limitation. The approach proposed in this paper requires careful selection of kernel width for Laplacian approximation, temperature parameters in the contrastive loss, and the learning rate for dynamic weights. Although we provide a scheme for hyperparameter tuning, improper parameters can lead to ineffective training. 

\section{Conclusion}\label{sec: conclusion}
This paper presents a method for structure-preserving contrastive learning for spatial time series, where we propose a dynamic mechanism to adaptively balance contrastive learning and structure preservation. The method is generally applicable to time series data with spatial or geographical features, which are particularly abundant in transportation systems. Extensive experiments demonstrate that our methods improve the state of the arts, including for multivariate time series classification in various contexts and for traffic prediction at both macroscopic and microscopic scales. 

An important advantage of this method is its ability to enforce the preservation of similarity structures in the latent space, thereby aligning representation learning with the original data manifold. At the same time, the efficiency costs of structure-preserving regularisation are moderate, making the approach practical for many applications. Nonetheless, several limitations deserve mention. The benefits of the method become expensive when time series are extremely long, where computational costs can be prohibitive. Dynamic weighting does not always outperform fixed weighting, but is particularly helpful when the training dynamics are dominated by one loss component. Moreover, the method depends on careful hyperparameter tuning, especially for the graph-geometry-preserving regulariser. These limitations do not undermine the contributions of the work but rather highlight opportunities for future research, including the design of lightweight approximations for long sequences, more stable dynamic weighting schemes, and principled strategies for automatic hyperparameter selection.

Notably, preserving the original similarity structure in the latent space is shown to be well aligned with and beneficial to contrastive learning for spatio-temporal data. This is crucial for informative representational learning from large-scale data, as it impacts a neural network’s ability to model the underlying conditional distribution. Our experiments suggest that higher similarity structure preservation is a good indicator of more informative representations, highlighting that the structural information of similarities in spatio-temporal data remains yet to be exploited. In the context of increasingly large neural network models that involve diverse data modalities, we hope this study sheds light on more effective training of large models in transportation systems.

\section*{Acknowledgments}
This work is supported by the TU Delft AI Labs programme. We extend our sincere gratitude to the researchers and organisations who collected, created, cleaned, and curated the high-quality datasets for research use, among which the UEA Multivariate Time Series Classification Archive is openly accessible at \url{https://www.timeseriesclassification.com/dataset.php}. We also would like to thank the anonymous reviewers for their valuable comments and advice.

The authors acknowledge the use of computational resources of the DelftBlue supercomputer, provided by Delft High Performance Computing Centre (https://www.tudelft.nl/dhpc).

\section*{Declaration of generative AI and AI-assisted technologies in the writing process}
During the preparation of this work the authors used ChatGPT and DeepSeek in order to obtain suggestions for readability improvement. No sentence was entirely generated by the generative tools. After using the tools, the authors have reviewed and edited the content as needed and take full responsibility for the content of the publication.


\bibliographystyle{elsarticle-num} 
\bibliography{references}







\newpage
\appendix
\renewcommand\thefigure{\thesection.\arabic{figure}}
\renewcommand{\theequation}{\thesection.\arabic{equation}}

\section{Detailed results on UEA datasets}\label{sec: apdx_UEA_details}
\setcounter{figure}{0}
\renewcommand{\thefigure}{A.\arabic{figure}}
\setcounter{equation}{0}
\renewcommand{\theequation}{A.\arabic{equation}}
\setcounter{table}{0}
\renewcommand{\thetable}{A.\arabic{table}}
This section provides detailed comparisons of evaluation results for the used 28 datasets in the UEA archive. Tables \ref{tab: uea_spatial_details} and \ref{tab: uea_nonspatial_details} present the results of classification accuracy. Tables \ref{tab: uea_spatial_details_time} and \ref{tab: uea_nonspatial_details_time} present the training time for self-supervised representation learning. 
\begin{table}[htb]
\caption{Detailed evaluation of classification accuracy on spatial datasets in the UEA archive.}
\label{tab: uea_spatial_details}
\begin{center}
\resizebox{\textwidth}{!}{%
\begin{tabular}{@{}lcccccccc@{}}
\toprule
Dataset & TS2Vec & Topo-TS2Vec & GGeo-TS2Vec & SoftCLT & Topo-SoftCLT & GGeo-SoftCLT \\ \midrule
ArticularyWordRecognition & 0.980 & \textbf{\underline{0.987}} & 0.983 & \textbf{\underline{0.987}} & 0.977 & \textbf{\underline{0.987}} \\
BasicMotions & \textbf{\underline{1.000}} & \textbf{\underline{1.000}} & \textbf{\underline{1.000}} & \textbf{\underline{1.000}} & \textbf{\underline{1.000}} & \textbf{\underline{1.000}} \\
CharacterTrajectories & 0.971 & 0.985 & 0.972 & 0.980 & 0.977 & \textbf{\underline{0.986}} \\
Cricket & 0.944 & 0.944 & 0.972 & 0.972 & 0.972 & \textbf{\underline{0.986}} \\
ERing & 0.867 & 0.874 & 0.881 & \textbf{\underline{0.893}} & 0.878 & 0.863 \\
EigenWorms & 0.809 & 0.817 & 0.863 & 0.817 & \textbf{\underline{0.901}} & 0.840 \\
Epilepsy & 0.957 & 0.957 & 0.949 & \textbf{\underline{0.964}} & 0.957 & 0.949 \\
Handwriting & 0.498 & 0.499 & 0.479 & 0.487 & 0.478 & \textbf{\underline{0.580}} \\
LSST & 0.485 & 0.566 & 0.536 & 0.452 & 0.569 & \textbf{\underline{0.581}} \\
Libras & 0.883 & 0.844 & 0.850 & \textbf{\underline{0.889}} & 0.850 & 0.867 \\
NATOPS & 0.917 & 0.917 & 0.933 & 0.922 & 0.917 & \textbf{\underline{0.944}} \\
PEMS-SF & 0.792 & 0.775 & \textbf{\underline{0.815}} & 0.751 & 0.803 & 0.740 \\
RacketSports & 0.908 & 0.914 & 0.914 & \textbf{\underline{0.928}} & 0.908 & 0.875 \\
UWaveGestureLibrary & 0.862 & 0.831 & 0.834 & 0.888 & 0.881 & \textbf{\underline{0.897}} \\ \midrule
Avg. over spatial datasets & 0.848 & 0.851 & 0.856 & 0.852 & 0.862 & \textbf{\underline{0.864}} \\ 
 \bottomrule
\end{tabular}}
\end{center}
\end{table}

\begin{table}[htb]
\caption{Detailed evaluation of classification accuracy on non-spatial datasets in the UEA archive.}
\label{tab: uea_nonspatial_details}
\begin{center}
\resizebox{\textwidth}{!}{%
\begin{tabular}{@{}lcccccccc@{}}
\toprule
Dataset & TS2Vec & Topo-TS2Vec & GGeo-TS2Vec & SoftCLT & Topo-SoftCLT & GGeo-SoftCLT \\ \midrule
AtrialFibrillation & 0.200 & \textbf{\underline{0.267}} & 0.133 & 0.133 & 0.200 & \textbf{\underline{0.267}} \\
DuckDuckGeese & 0.360 & \textbf{\underline{0.540}} & 0.520 & 0.400 & 0.420 & 0.400 \\
EthanolConcentration & 0.289 & 0.274 & 0.297 & 0.243 & \textbf{\underline{0.308}} & \textbf{\underline{0.308}} \\
FaceDetection & 0.510 & 0.508 & 0.505 & \textbf{\underline{0.516}} & 0.497 & 0.505 \\
FingerMovements & 0.480 & 0.480 & 0.480 & 0.530 & 0.470 & \textbf{\underline{0.540}} \\
HandMovementDirection & 0.324 & \textbf{\underline{0.405}} & 0.257 & 0.324 & 0.230 & 0.257 \\
Heartbeat & 0.751 & \textbf{\underline{0.761}} & 0.717 & 0.756 & 0.737 & 0.732 \\
JapaneseVowels & 0.978 & \textbf{\underline{0.986}} & 0.978 & 0.970 & 0.978 & 0.978 \\
MotorImagery & 0.480 & 0.500 & 0.500 & \textbf{\underline{0.520}} & 0.500 & 0.500 \\
PhonemeSpectra & 0.263 & 0.258 & \textbf{\underline{0.269}} & \textbf{\underline{0.269}} & 0.260 & 0.257 \\
SelfRegulationSCP1 & 0.778 & 0.768 & \textbf{\underline{0.788}} & 0.761 & 0.730 & 0.771 \\
SelfRegulationSCP2 & 0.467 & 0.550 & \textbf{\underline{0.561}} & 0.528 & 0.511 & 0.511 \\
SpokenArabicDigits & 0.973 & \textbf{\underline{0.976}} & 0.966 & 0.964 & 0.968 & 0.957 \\
StandWalkJump & 0.467 & 0.467 & \textbf{\underline{0.533}} & 0.200 & 0.133 & \textbf{\underline{0.533}} \\ \midrule
Avg. over non-spatial datasets & 0.523 & \textbf{\underline{0.553}} & 0.536 & 0.508 & 0.496 & 0.537 \\ 
 \bottomrule
\end{tabular}}
\end{center}
\end{table}

\begin{table}[htb]
\caption{Detailed representation training time per epoch (unit: s) on spatial datasets in the UEA archive.}
\label{tab: uea_spatial_details_time}
\begin{center}
\resizebox{\textwidth}{!}{%
\begin{tabular}{@{}lcccccccc@{}}
\toprule
Dataset & TS2Vec & Topo-TS2Vec & GGeo-TS2Vec & SoftCLT & Topo-SoftCLT & GGeo-SoftCLT \\ \midrule
ArticularyWordRecognition & 3.799 (1.00$\times$) & 5.61 (1.48$\times$) & 5.863 (1.54$\times$) & 3.772 (0.99$\times$) & 5.77 (1.52$\times$) & 5.983 (1.57$\times$) \\
BasicMotions & 0.475 (1.00$\times$) & 0.685 (1.44$\times$) & 0.709 (1.49$\times$) & 0.457 (0.96$\times$) & 0.687 (1.45$\times$) & 0.711 (1.50$\times$) \\
CharacterTrajectories & 20.640 (1.00$\times$) & 30.863 (1.50$\times$) & 33.32 (1.61$\times$) & 20.652 (1.00$\times$) & 30.948 (1.50$\times$) & 33.18 (1.61$\times$) \\
Cricket & 1.903 (1.00$\times$) & 2.653 (1.39$\times$) & 5.437 (2.86$\times$) & 1.904 (1.00$\times$) & 2.655 (1.40$\times$) & 5.436 (2.86$\times$) \\
ERing & 0.319 (1.00$\times$) & 0.482 (1.51$\times$) & 0.487 (1.53$\times$) & 0.316 (0.99$\times$) & 0.483 (1.51$\times$) & 0.49 (1.54$\times$) \\
EigenWorms & 19.862 (1.00$\times$) & 23.823 (1.20$\times$) & 149.05 (7.50$\times$) & 20.224 (1.02$\times$) & 24.856 (1.25$\times$) & 150.7 (7.59$\times$) \\
Epilepsy & 1.737 (1.00$\times$) & 2.49 (1.43$\times$) & 2.753 (1.58$\times$) & 1.686 (0.97$\times$) & 2.506 (1.44$\times$) & 2.755 (1.59$\times$) \\
Handwriting & 1.875 (1.00$\times$) & 2.771 (1.48$\times$) & 2.959 (1.58$\times$) & 1.88 (1.00$\times$) & 2.775 (1.48$\times$) & 2.987 (1.59$\times$) \\
LSST & 29.786 (1.00$\times$) & 45.273 (1.52$\times$) & 45.162 (1.52$\times$) & 29.859 (1.00$\times$) & 45.216 (1.52$\times$) & 45.154 (1.52$\times$) \\
Libras & 2.081 (1.00$\times$) & 3.142 (1.51$\times$) & 3.142 (1.51$\times$) & 2.085 (1.00$\times$) & 3.135 (1.51$\times$) & 3.141 (1.51$\times$) \\
NATOPS & 1.953 (1.00$\times$) & 2.989 (1.53$\times$) & 2.949 (1.51$\times$) & 2.085 (1.07$\times$) & 3.147 (1.61$\times$) & 3.159 (1.62$\times$) \\
PEMS-SF & 3.413 (1.00$\times$) & 5.069 (1.49$\times$) & 5.38 (1.58$\times$) & 3.415 (1.00$\times$) & 5.064 (1.48$\times$) & 5.399 (1.58$\times$) \\
RacketSports & 1.781 (1.00$\times$) & 2.685 (1.51$\times$) & 2.664 (1.50$\times$) & 1.771 (0.99$\times$) & 2.711 (1.52$\times$) & 2.665 (1.50$\times$) \\
UWaveGestureLibrary & 1.699 (1.00$\times$) & 2.395 (1.41$\times$) & 2.788 (1.64$\times$) & 1.776 (1.05$\times$) & 2.595 (1.53$\times$) & 2.99 (1.76$\times$) \\ \midrule
Avg. over spatial datasets & 6.523 sec/epoch & 1.46$\times$ & 2.12$\times$ & 1.00$\times$ & 1.48$\times$ & 2.15$\times$ \\
\bottomrule
\end{tabular}}
\end{center}
\end{table}

\begin{table}[htb]
\caption{Detailed representation training time per epoch (unit: s) on non-spatial datasets in the UEA archive.}
\label{tab: uea_nonspatial_details_time}
\begin{center}
\resizebox{\textwidth}{!}{%
\begin{tabular}{@{}lcccccccc@{}}
\toprule
Dataset & TS2Vec & Topo-TS2Vec & GGeo-TS2Vec & SoftCLT & Topo-SoftCLT & GGeo-SoftCLT \\ \midrule
AtrialFibrillation & 0.182 (1.00$\times$) & 0.258 (1.42$\times$) & 0.369 (2.03$\times$) & 0.177 (0.97$\times$) & 0.259 (1.42$\times$) & 0.366 (2.01$\times$) \\
DuckDuckGeese & 0.617 (1.00$\times$) & 0.973 (1.58$\times$) & 1.059 (1.72$\times$) & 0.621 (1.01$\times$) & 0.968 (1.57$\times$) & 1.104 (1.79$\times$) \\
EthanolConcentration & 4.939 (1.00$\times$) & 6.655 (1.35$\times$) & 20.128 (4.08$\times$) & 4.89 (0.99$\times$) & 6.664 (1.35$\times$) & 20.182 (4.09$\times$) \\
FaceDetection & 70.709 (1.00$\times$) & 109.6 (1.55$\times$) & 108.83 (1.54$\times$) & 71.104 (1.01$\times$) & 107.523 (1.52$\times$) & 107.092 (1.51$\times$) \\
FingerMovements & 3.826 (1.00$\times$) & 5.67 (1.48$\times$) & 5.706 (1.49$\times$) & 3.779 (0.99$\times$) & 5.671 (1.48$\times$) & 5.716 (1.49$\times$) \\
HandMovementDirection & 2.221 (1.00$\times$) & 3.353 (1.51$\times$) & 4.151 (1.87$\times$) & 2.226 (1.00$\times$) & 3.334 (1.50$\times$) & 4.142 (1.86$\times$) \\
Heartbeat & 2.811 (1.00$\times$) & 4.218 (1.50$\times$) & 5.22 (1.86$\times$) & 2.818 (1.00$\times$) & 4.216 (1.50$\times$) & 5.221 (1.86$\times$) \\
JapaneseVowels & 3.211 (1.00$\times$) & 4.871 (1.52$\times$) & 4.821 (1.50$\times$) & 3.199 (1.00$\times$) & 4.846 (1.51$\times$) & 4.83 (1.50$\times$) \\
MotorImagery & 7.450 (1.00$\times$) & 9.637 (1.29$\times$) & 51.0 (6.85$\times$) & 7.475 (1.00$\times$) & 9.659 (1.30$\times$) & 50.881 (6.83$\times$) \\
PhonemeSpectra & 42.956 (1.00$\times$) & 63.801 (1.49$\times$) & 70.578 (1.64$\times$) & 43.015 (1.00$\times$) & 63.807 (1.49$\times$) & 70.806 (1.65$\times$) \\
SelfRegulationSCP1 & 4.178 (1.00$\times$) & 6.042 (1.45$\times$) & 10.446 (2.50$\times$) & 4.237 (1.01$\times$) & 6.094 (1.46$\times$) & 10.415 (2.49$\times$) \\
SelfRegulationSCP2 & 3.295 (1.00$\times$) & 4.67 (1.42$\times$) & 9.391 (2.85$\times$) & 3.269 (0.99$\times$) & 4.629 (1.40$\times$) & 9.376 (2.85$\times$) \\
SpokenArabicDigits & 96.299 (1.00$\times$) & 143.411 (1.49$\times$) & 131.577 (1.37$\times$) & 86.495 (0.90$\times$) & 125.916 (1.31$\times$) & 129.073 (1.34$\times$) \\
StandWalkJump & 0.304 (1.00$\times$) & 0.404 (1.33$\times$) & 1.68 (5.53$\times$) & 0.31 (1.02$\times$) & 0.400 (1.32$\times$) & 1.7 (5.59$\times$) \\ \midrule
Avg. over non-spatial datasets & 17.357 sec/epoch & 1.46$\times$ & 2.57$\times$ & 0.99$\times$ & 1.44$\times$ & 2.57$\times$ \\
\bottomrule
\end{tabular}}
\end{center}
\end{table}

\begin{figure}[htb]
\centering
\begin{center}
    \includegraphics[width=\textwidth]{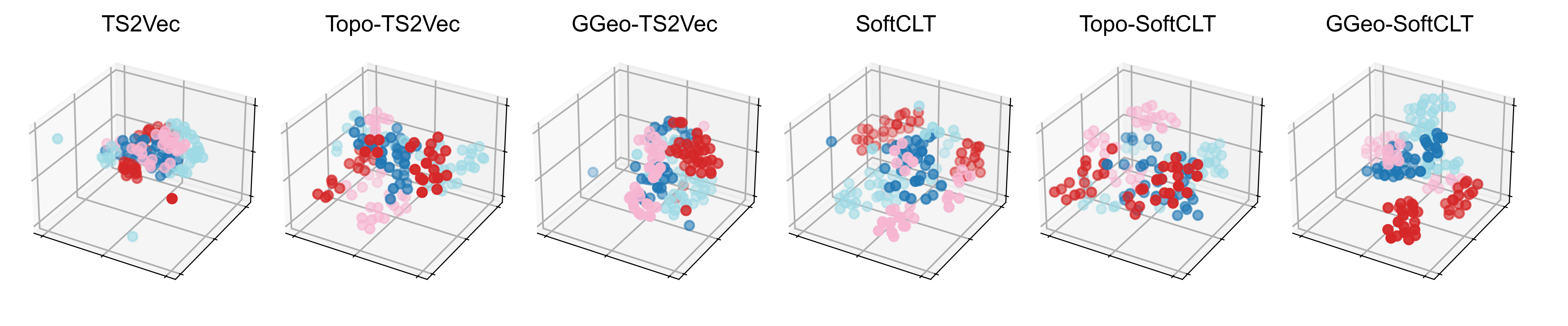}
\end{center}
\caption{Encoded representations after training with different losses on the test set of Epilepsy.}
\label{fig: latent_epilepsy}
\begin{center}
    \includegraphics[width=\textwidth]{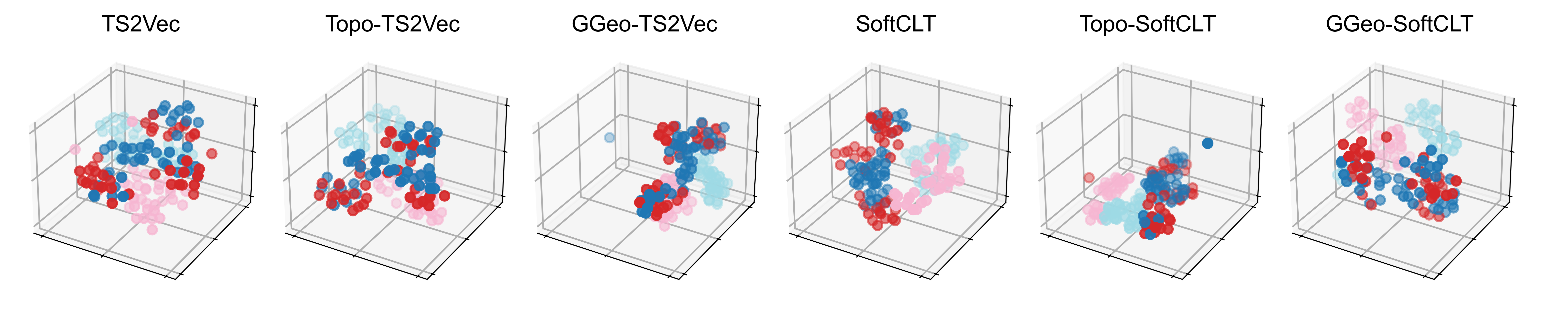}
\end{center}
\caption{Encoded representations after training with different losses on the test set of RacketSports.}
\label{fig: latent_racketsports}
\end{figure}

In addition, to visually show the effect of differently regularised contrastive learning losses on representation, we apply t-SNE to compress the encoded representations into 3 dimensions, as plotted in Figure \ref{fig: latent_epilepsy} for the dataset Epilepsy, and Figure \ref{fig: latent_racketsports} for RacketSports. The classes are indicated by colours. We use these two datasets because they are visualisation-friendly, with 4 classes and around 150 test samples. 

\section{Detailed results of macroscopic prediction with LSTM and GRU}\label{sec: apdx_lstm_gru_details}
\setcounter{figure}{0}
\renewcommand{\thefigure}{B.\arabic{figure}}
\setcounter{equation}{0}
\renewcommand{\theequation}{B.\arabic{equation}}
\setcounter{table}{0}
\renewcommand{\thetable}{B.\arabic{table}}
This section provides additional tables and figures presenting the evaluation of the final results using LSTM and GRU in macroscopic traffic prediction. Table \ref{tab: lstm_gru_prediction} shows the task-specific metrics and Table \ref{tab: lstm_gru_structure} shows the metrics for global structure preservation. Figures \ref{fig: latent_lstm} and \ref{fig: latent_gru} show the latent representations encoded by LSTM and GRU models, respectively.

\begin{table}[htb]
\caption{Macroscopic traffic prediction evaluation with LSTM and GRU encoders. Metrics are reported as mean $\pm$ standard deviation over 10 stratified folds of the test set.}
\label{tab: lstm_gru_prediction}
\begin{center}
\resizebox{\textwidth}{!}{%
\begin{tabular}{@{}lcccccccc@{}}
\toprule
\multicolumn{1}{c}{\multirow{2}{*}{Method}} & \multicolumn{4}{c}{LSTM} & \multicolumn{4}{c}{GRU} \\ \cmidrule(l){2-9} 
\multicolumn{1}{c}{} & \begin{tabular}[c]{@{}c@{}}MAE (↓)\\ (km/h)\end{tabular} & \begin{tabular}[c]{@{}c@{}}RMSE (↓)\\ (km/h)\end{tabular} & \begin{tabular}[c]{@{}c@{}}SDEP (↓)\\ (km/h)\end{tabular} & \multicolumn{1}{c}{\begin{tabular}[c]{@{}c@{}}EVar (↑)\\ (\%)\end{tabular}} & \begin{tabular}[c]{@{}c@{}}MAE (↓)\\ (km/h)\end{tabular} & \begin{tabular}[c]{@{}c@{}}RMSE (↓)\\ (km/h)\end{tabular} & \begin{tabular}[c]{@{}c@{}}SDEP (↓)\\ (km/h)\end{tabular} & \begin{tabular}[c]{@{}c@{}}EVar (↑)\\ (\%)\end{tabular} \\ \midrule
No pretraining & 3.244 $\pm$ 0.046 & 6.401 $\pm$ 0.102 & 6.399 $\pm$ 0.102 & 82.151 $\pm$ 0.512 & 3.552 $\pm$ 0.063 & 7.227 $\pm$ 0.119 & 7.227 $\pm$ 0.119 & 77.241 $\pm$ 0.457 \\
TS2Vec & 3.158 $\pm$ 0.045 & 6.187 $\pm$ 0.089 & 6.186 $\pm$ 0.089 & 83.322 $\pm$ 0.413 & 3.601 $\pm$ 0.062 & 7.296 $\pm$ 0.113 & 7.296 $\pm$ 0.113 & 76.805 $\pm$ 0.427 \\
Topo-TS2Vec & 3.139 $\pm$ 0.048 & 6.154 $\pm$ 0.107 & 6.153 $\pm$ 0.107 & 83.499 $\pm$ 0.423 & 3.491 $\pm$ 0.057 & 7.005 $\pm$ 0.108 & 7.005 $\pm$ 0.108 & 78.617 $\pm$ 0.442 \\
GGeo-TS2Vec & \textbf{\underline{3.101}} $\pm$ 0.046 & \textbf{\underline{5.974}} $\pm$ 0.089 & \textbf{\underline{5.973}} $\pm$ 0.089 & \textbf{\underline{84.454}} $\pm$ 0.354 & 3.466 $\pm$ 0.055 & 6.947 $\pm$ 0.105 & 6.946 $\pm$ 0.105 & 78.975 $\pm$ 0.417 \\
SoftCLT & 3.191 $\pm$ 0.049 & 6.319 $\pm$ 0.100 & 6.318 $\pm$ 0.100 & 82.604 $\pm$ 0.412 & \textbf{3.349} $\pm$ 0.051 & \textbf{6.649} $\pm$ 0.101 & \textbf{6.648} $\pm$ 0.101 & \textbf{80.741} $\pm$ 0.388 \\
Topo-SoftCLT & 3.192 $\pm$ 0.049 & 6.270 $\pm$ 0.101 & 6.270 $\pm$ 0.101 & 82.872 $\pm$ 0.317 & 3.367 $\pm$ 0.049 & 6.660 $\pm$ 0.098 & 6.659 $\pm$ 0.098 & 80.674 $\pm$ 0.398 \\
GGeo-SoftCLT & \textbf{3.128} $\pm$ 0.044 & \textbf{6.121} $\pm$ 0.094 & \textbf{6.120} $\pm$ 0.094 & \textbf{83.675} $\pm$ 0.400 & \textbf{\underline{3.302}} $\pm$ 0.049 & \textbf{\underline{6.494}} $\pm$ 0.097 & \textbf{\underline{6.494}} $\pm$ 0.097 & \textbf{\underline{81.621}} $\pm$ 0.430 \\
\midrule
Best improvement & 4.415 & 6.675 & 6.667 & 2.803 & 7.026 & 10.139 & 10.143 & 5.671 \\
\bottomrule
\multicolumn{9}{l}{Note: the \textbf{\underline{best}} values are both bold and underlined; the \textbf{second-best} values are bold.}
\end{tabular}}
\end{center}
\end{table}

\begin{table}[htb]
\caption{Global structure preservation of LSTM and GRU encoders in macroscopic traffic prediction task.}
\label{tab: lstm_gru_structure}
\begin{center}
\resizebox{\textwidth}{!}{%
\begin{tabular}{@{}lcccccccccc@{}}
\toprule
\multicolumn{1}{c}{\multirow{2}{*}{Method}} & \multicolumn{5}{c}{LSTM} & \multicolumn{5}{c}{GRU} \\ \cmidrule(l){2-11} 
\multicolumn{1}{c}{} & kNN (↑) & Trust. (↑) & Cont. (↑) & MRRE (↓) & dRMSE (↓) & kNN (↑) & Trust. (↑) & Cont. (↑) & MRRE (↓) & dRMSE (↓) \\ \midrule
No pretraining & \textbf{\underline{0.215}} & 0.929 & 0.942 & 0.046 & 0.319 & 0.207 & 0.946 & 0.951 & 0.037 & \textbf{0.299} \\
TS2Vec & 0.173 & \textbf{0.959} & \textbf{0.955} & \textbf{0.035} & \textbf{0.318} & 0.197 & \textbf{0.963} & 0.957 & 0.031 & \textbf{\underline{0.269}} \\
Topo-TS2Vec & 0.179 & 0.947 & 0.945 & 0.042 & 0.332 & 0.193 & \textbf{0.963} & \textbf{0.962} & 0.030 & 0.304 \\
GGeo-TS2Vec & \textbf{0.198} & \textbf{\underline{0.968}} & \textbf{\underline{0.961}} & \textbf{\underline{0.029}} & \textbf{\underline{0.299}} & 0.203 & 0.962 & \textbf{0.962} & \textbf{0.029} & 0.306 \\
SoftCLT & 0.186 & 0.954 & 0.948 & 0.038 & 0.334 & 0.214 & 0.958 & 0.957 & 0.031 & 0.319 \\
Topo-SoftCLT & 0.193 & 0.947 & 0.950 & 0.039 & 0.326 & \textbf{\underline{0.239}} & \textbf{\underline{0.966}} & \textbf{\underline{0.964}} & \textbf{\underline{0.026}} & 0.354 \\
GGeo-SoftCLT & 0.176 & 0.945 & 0.942 & 0.042 & 0.365 & \textbf{0.228} & \textbf{0.963} & 0.957 & \textbf{0.029} & 0.339 \\
\bottomrule
\multicolumn{11}{l}{Note: the \textbf{\underline{best}} values are both bold and underlined; the \textbf{second-best} values are bold.}
\end{tabular}}
\end{center}
\end{table}

\begin{figure}[htb]
\centering
\begin{center}
    \includegraphics[width=\textwidth]{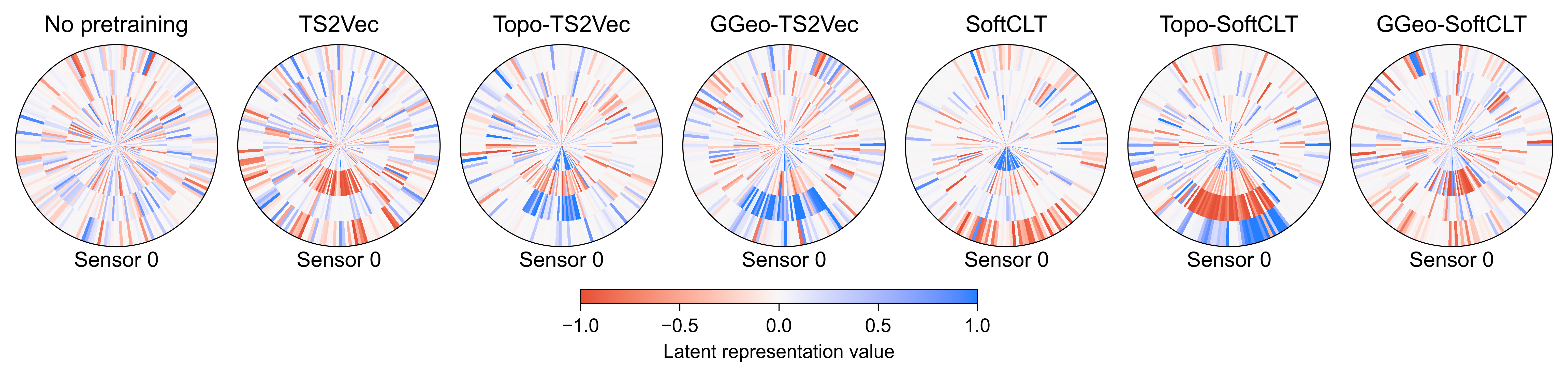}
\end{center}
\caption{LSTM encoded representations after training with different losses on the test set of the Macroscopic traffic prediction task.}
\label{fig: latent_lstm}
\begin{center}
    \includegraphics[width=\textwidth]{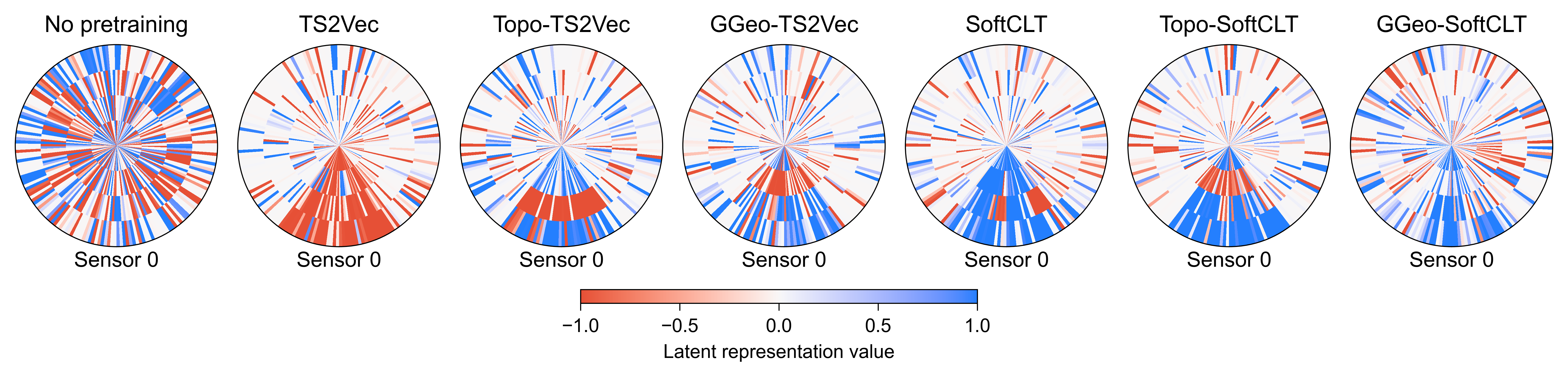}
\end{center}
\caption{GRU encoded representations after training with different losses on the test set of the Macroscopic traffic prediction task.}
\label{fig: latent_gru}
\end{figure}

\end{document}